\documentclass[a4paper,fleqn]{cas-sc}

\usepackage[authoryear]{natbib}
\bibpunct{(}{)}{;}{a}{}{,}

\usepackage{amsmath}
\usepackage{graphicx}
\usepackage{subcaption}
\usepackage{float}
\usepackage{placeins}
\usepackage{lipsum}
\usepackage{soul,color}
\usepackage{subcaption}
\usepackage{algorithm}
\usepackage{algpseudocode}
\usepackage{subcaption}
\usepackage{lineno}
\def\tsc#1{\csdef{#1}{\textsc{\lowercase{#1}}\xspace}}
\tsc{WGM}
\tsc{QE}
\tsc{EP}
\tsc{PMS}
\tsc{BEC}
\tsc{DE}

\begin{document}

\let\WriteBookmarks\relax
\def\floatpagepagefraction{1}
\def\textpagefraction{.001}
\shorttitle{A Real-time Evaluation Framework for Pedestrian's Potential Risk at Non-Signalized Intersections Based on Predicted Post-Encroachment Time}
\shortauthors{T. Lin et al.}
%\begin{frontmatter}

\title [mode = title]{A Real-time Evaluation Framework for Pedestrian's Potential Risk at Non-Signalized Intersections Based on Predicted Post-Encroachment Time}

% \tnotemark[1,2]

% \tnotetext[1]{This research was supported by Basic Science Research Program through the National Research Foundation of Korea(NRF) funded by the Ministry of  Science and ICT(NRF-2017R1A2B2002329)}

% \tnotetext[2]{The second title footnote which is a longer text matter
%   to fill through the whole text width and overflow into
%   another line in the footnotes area of the first page.}

\author[1]{Tengfeng Lin}[orcid=0000-0003-1397-6603]
% \fnmark[1]
% \cormark[1]
\ead{tengfenglin@kaist.ac.kr}
\credit{Conceptualization,
Methodology,
Data curation,
Software,
Validation,
Visualization,
Formal analysis,
Writing - Original Draft}

\author[2, 3]{Zhixiong Jin}[orcid=0000-0002-1370-781X]
\ead{zhixiong.jin@univ-eiffel.fr}
\credit{Conceptualization, Methodology, Data curation, Writing - review \& editing}
\author[4]{Seongjin Choi}[orcid=0000-0001-7140-537X]
% \fnmark[1]
\ead{chois@umn.edu}
\credit{Conceptualization, Methodology, Writing - review \& editing}

\author[5]{Hwasoo Yeo}[orcid=0000-0002-2684-0978]
% \fnmark[1]
\cormark[1]
\ead{hwasoo@kaist.ac.kr}
\credit{Conceptualization, Data curation, Resources,
Supervision, Funding acquisition, Project administration, Writing - review \& editing}

\address[1]{Cho Chun Shik Graduate School of Mobility, Korea Advanced Institute of Science and Technology, Daejeon, Republic of Korea}

\address[2]{Univ. Guatave Eiffel, ENTPE, LICIT-ECO7, Lyon, France}
\address[3]{Urban Transport Systems Laboratory (LUTS), École Polytechnique Fédérale de Lausanne (EPFL), Lausanne, CH 1015, Switzerland}
\address[4]{Department of Civil, Environmental, and Geo- Engineering, University of Minnesota, 500 Pillsbury Dr. SE, Minneapolis, MN 55455, USA}
\address[5]{Department of Civil and Environmental Engineering, Korea Advanced Institute of Science and Technology, 291 Daehak-ro, Yuseong-gu, Daejeon, 34141, Republic of Korea}

\cortext[cor1]{Corresponding author}

\begin{highlights}
\item A framework with computer vision technologies and predictive models is proposed to evaluate pedestrian's potential risk in real time.
\item A novel predicted surrogate safety measure, Predicted Post-Encroachment Time (P-PET), is introduced to accurately and interpretably evaluate potential pedestrian risk.
\item Pedestrians are classified into kids, adults, and cyclists to tailor specific evaluation rules to each category of pedestrians, thus enhancing the effectiveness and reliability of risk evaluation.
\item The results at a real-field, non-signalized intersection show the effectiveness of the proposed framework and illustrate the computational time costs across different components, demonstrating its potential in real-time application.

\end{highlights}

\begin{abstract}
Addressing pedestrian safety at intersections is one of the paramount concerns in the field of transportation research, driven by the urgency of reducing traffic-related injuries and fatalities. With advances in computer vision technologies and predictive models, the pursuit of developing real-time proactive protection systems is increasingly recognized as vital to improving pedestrian safety at intersections. The core of these protection systems lies in the prediction-based evaluation of pedestrian's potential risks, which plays a significant role in preventing the occurrence of accidents. The major challenges in the current prediction-based potential risk evaluation research can be summarized into three aspects: the inadequate progress in creating a real-time framework for the evaluation of pedestrian's potential risks, the absence of accurate and explainable safety indicators that can represent the potential risk, and the lack of tailor-made evaluation criteria specifically for each category of pedestrians. 
To address these research challenges, in this study, a framework with computer vision technologies and predictive models is developed to evaluate the potential risk of pedestrians in real time. Integral to this framework is a novel surrogate safety measure, the Predicted Post-Encroachment Time (P-PET), derived from deep learning models capable to predict the arrival time of pedestrians and vehicles at intersections. To further improve the effectiveness and reliability of pedestrian risk evaluation,  we classify pedestrians into distinct categories and apply specific evaluation criteria for each group.
The experiment is conducted at a real field non-signalized crosswalk.
The results demonstrate the framework's ability to effectively identify potential risks through the use of P-PET, indicating its feasibility for real-time applications and its improved performance in risk evaluation across different categories of pedestrians.
\end{abstract}

\begin{keywords}
Computer Vision\sep
Proactive Safety Protection\sep
Surrogate Safety Measures\sep
Pedestrians Risk Evaluation\sep 
Arrival-time Prediction\sep 
\end{keywords}

\maketitle

\section{Introduction}
\subsection{Background}
Pedestrian safety is becoming more urgent in transportation research, with pedestrians accounting for 17\% of 35,000 traffic deaths in the United States in 2019 and more than 76,000 injuries in 2020 \citep{stewart2022overview}.
%Intersecition is critical
In particular, a substantial proportion of fatal traffic accidents occurred at intersections \citep{who2013global, lawstraffic, TransportCanada2021}. Intersections pose a higher risk to pedestrians compared to other roads due to the lack of physical barriers that separate them from vehicles.
%intersection safety is important
Recently, with advances in computer vision technologies and predictive models, the pursuit of developing real-time proactive protection systems is increasingly recognized as vital to improving pedestrian safety at intersections. These systems require methods to evaluate the potential risk for pedestrians from approaching vehicles, underlining that understanding how to evaluate pedestrian risk is an indispensable aspect of pedestrian protection.

\subsection{Review of Current Research}
\subsubsection{Traditional Pedestrian Safety Analysis}

Numerous studies have evaluated pedestrian risk by examining the various factors that influence both the occurrence \citep{gaarder2004impact, wong2007contributory, wier2009area, wang2016macro, lee2005comprehensive} and the severity \citep{batouli2020analysis, olszewski2015pedestrian,choi2013pedestrian} of pedestrian-vehicle crashes at intersections according to the crash dataset. In particular, \cite{gaarder2004impact} investigated pedestrian crashes using a dataset of 1,589 reported incidents from 1994 to 1998, focusing on how speeding influences crash rates. 
\cite{wong2007contributory} analyzed a crash dataset from 262 intersections spanning 2002 to 2003, using Poisson and negative binomial regressions to pinpoint significant risk factors such as road environment, curvature and tram stops. \cite{wier2009area} analyzed 4,039 crash events, identifying traffic volume as a critical environmental factor that influences the occurrence of vehicle-pedestrian injury collisions. Regarding the severity of pedestrian-vehicle collisions, \cite{batouli2020analysis} explored the severity of pedestrian outcomes of motor vehicle crashes in Colorado, identifying key factors such as intersection proximity, lighting conditions, and characteristics of pedestrians and drivers as significant. \cite{olszewski2015pedestrian} then applied a binary logit model to assess pedestrian safety in Poland, revealing the impact of environmental conditions, road types, and demographic factors on the severity of the collision. \cite{choi2013pedestrian} utilized logistic regression to identify additional characteristics that influenced severity, such as night, weather, and location of the crash. 

In summary, these researches pinpointing key factors that influence the occurrence and severity of pedestrian-vehicle collisions at intersections suggest that post-collision interventions, such as curb bulb installations and traffic volume restrictions, are crucial to mitigate further risks. 
However, these studies face significant obstacles in obtaining high-quality large-scale collision data, sourced primarily from police reports or insurance claims, and ethical concerns related to waiting for crash occurrences while trying to prevent such incidents.

\subsubsection{Vision techniques in pedestrian trajectory extraction}
Driven by rapid advances in computer vision technologies and the proliferation of data from vision sensors deployed on urban road networks, the study of pedestrian safety at intersections is entering a new era. This progress allows us to extract spatial and temporal information of road users at intersections. These data enable a more precise and effective analysis of potential conflicts and risks, offering a new perspective on overcoming the challenges identified in previous studies. Therefore, numerous studies use advanced detection and tracking technologies to collect data from vision sensors to evaluate the probability and severity of crashes.
These studies can be summarized into two classes: post-interaction-based \citep{ni2016evaluation, chen2017surrogate,fu2016pedestrian, wu2020improved,vasudevan2022lidar} and prediction-based \citep{lim2018real,kathuria2020evaluating,zhang2020modeling,zhang2020prediction,baek2020vehicle,noh2022novel}.

\subsubsection{Post-interaction-based pedestrian risk evaluation study}
The post-interaction-based study focuses on analyzing complete trajectory data captured from vision sensors to assess pedestrian risks. These studies emphasize the use of Surrogate Safety Measures, which are crucial to evaluate pedestrian-vehicle conflicts in near-crash scenarios without an actual crash occurring. 
These measures, such as Post-Encroachment Time (PET) \citep{allen1978analysis}, Time to Collision (TTC) \citep{hayward1971near}, and Gap Time (GT) \citep{VOGEL200215}, are important in identifying potential conflicts at intersections. For example, \cite{ni2016evaluation} used the Traffic Analyzer Tool to extract the trajectories from the video dataset at three signalized intersections. By analyzing TTC and GT, three different patterns of pedestrian-vehicle interaction were identified, providing information on safety levels from a behavioral point of view. \cite{chen2017surrogate} used Faster R-CNN and Kemelized correlation filters to process the dataset captured by drones at a signalized intersection to extract trajectory information. This process used PET and relative time to collision (RTTC) to present pedestrian-vehicle conflicts both spatially and temporally.

Meanwhile, although LiDAR datasets may not provide the extensive information found in video datasets, some researchers prefer them for trajectory analysis due to their superior ability to detect pedestrians with overlaps. This is achieved by accurately mapping pedestrian positions in three dimensions, compensating for the lack of visual detail with precise spatial positioning.
\cite{wu2020improved} processed roadside LiDAR data using a sequence of computer vision techniques to extract the dataset of the trajectory of pedestrians and vehicles. Near-crash pedestrian-vehicle events were successfully identified and analyzed by applying three distinct surrogate safety measures (SSMs): PET, Proportion of Stopping Distance (PSD) and Crash Potential Index, utilizing the extracted dataset. Employing similar LiDAR data processing techniques, \cite{vasudevan2022lidar} used modified Post-encroachment Time based on a trajectory to analyze the behaviors of pedestrians at higher risk. Their study revealed that motorcycle-pedestrian conflicts are more risky compared to car-pedestrian conflicts. 
In essence, SSMs are indispensable in post-interaction analyzes, enabling the depiction of near-crash scenarios without the occurrence of actual crashes, thereby addressing data scarcity. 

However, the reliance on completed trajectory data collected after near-crash conflicts limits these studies. This requires that safety strategies and policies are formulated reactively, post-incident, highlighting a gap in the proactive evaluation of pedestrian risks and the application of preventive safety measures.

\subsubsection{Predicted-based potential risk evaluation study}
%## use portial trajectory of pedestrian to predict the future risk
Addressing the constraints of post-interaction analysis, research has shifted towards a predictive approach, evaluating potential pedestrian risks through the analysis of partial trajectories prior to conflict occurrences. This approach seeks to implement proactive safety measures, allowing early identification and evaluation of potential risks for pedestrians before conflicts arise\citep{kathuria2020evaluating,zhang2020modeling,zhang2020prediction,baek2020vehicle,noh2022novel}. 
%## have linear assumption
For example, \cite{kathuria2020evaluating} predicted future pedestrian-vehicle conflicts by analyzing trajectory data before the actual pedestrian crossing occurs at nonsignalized intersections. The average moving velocity obtained from the partial trajectory data was utilized to estimate the value of SSMs such as TTC, PET, and GT to indicate the potential conflict.
With a similar idea, \cite{baek2020vehicle} used Kalman filtering to process partial trajectory data, focusing on motion and spatial information before conflicts occurred, to predict future vehicle-to-vehicle conflicts. This method involves calculating the expected TTC values, employing TTC as an indicator to predict potential risks between vehicles.
However, the effectiveness of these methods in forecasting the future state of pedestrians and vehicles is constrained by the oversimplified linear assumption of pedestrian moving behavior, which is not valid in complex and unpredictable scenarios, limiting the accuracy of long-term predictions \citep{yang2009kalman}.
%## Deep Learning

On the contrary, deep learning models such as Gated Recurrent Unit (GRU) \citep{chung2014empirical}, Long-Short-Term Memory (LSTM) \citep{hochreiter1997long}, and Transformers \citep{vaswani2017attention} offer a non-linear data-driven approach to understanding data. Researchers are now utilizing these advanced models to more accurately predict future states from current information, overcoming the constraints of oversimplified assumptions.
%## Trajectory Prediction
For example, \cite{noh2022novel} estimate potential areas of collision risk using LSTM to predict trajectories. This work shows the feasibility of employing trajectory predictions to predict future interactions. 
However, it faces limitations in the increased computational cost and the lack of convincing safety evaluation metrics. Furthermore, while trajectory predictions maintain accuracy over short distances, precision decreases significantly at longer distances \citep{salzmann2020trajectron++}.
\cite{zhang2020modeling} and \cite{zhang2020prediction} investigated an alternative approach to assess pedestrian risk potential using deep learning models. Utilizing partial trajectory data preceding pedestrian crossings, they applied LSTM or GRU models for an end-to-end prediction of pedestrian future risk levels. These risk levels were quantified using the minimum Time To Collision (mTTC) and the Post-Encroachment Time (PET) with predefined threshold.
This work demonstrates the benefits of risk prediction using traditional SSMs, highlighting their advantages of simplification of complex interactions in the future and lower computational cost. 
However, these studies face several obstacles. Although their aim is to proactively protect pedestrians by predicting risks, there is a notable gap in proving the real-time applicability of these predicted-based methods. Furthermore, the approach of deriving the risk level from end-to-end prediction models reduces its explainability. Additionally, the reliance on traditional SSMs to determine risk still meets the limitations in conflict scenarios where pedestrian intentions play a critical role. This challenge is notably illustrated when traditional SSMs erroneously suggest safety in situations that have potential risks. For example, consider a scenario where a pedestrian decelerates due to an approaching vehicle, resulting in a high value of traditional SSMs. Despite traditional SSMs suggesting a safety scenario, the situation should, in fact, be recognized as having a risk of conflict. 

\subsection{Research limitation Summarizing}
In conclusion, the main research gaps in prediction-based potential risk evaluation studies can be summarized into three aspects. 
First, existing studies on predicted-based risk evaluation lack a framework for real-time application. Second, it is necessary to develop accurate and explainable safety indicators that can represent the pedestrian's potential risk. In addition, recognizing that vulnerable pedestrian groups, such as kids, demonstrate unique behavioral dynamics and face distinct levels of risk at intersections\citep{gitelman2019exploring,ferenchak2022residential,rothman2022child}, it is crucial to tailor evaluation criteria specifically for each category of pedestrians.

\subsection{Research contribution}
To address the identified research gaps and protect pedestrian safety proactively, a framework with computer vision technologies and predictive models is developed to evaluate the potential risk of pedestrians in real time. Integral to this framework is a novel surrogate safety measure, Predicted Post-Encroachment Time (P-PET), derived from deep learning models capable to predict the arrival time of pedestrians and vehicles at intersections. To further improve the effectiveness and reliability of the pedestrian risk evaluation, we classify pedestrians into three distinct categories, kids, adults, and cyclists, and apply specific evaluation criteria for each group. In this framework, computer vision is used to collect trajectory data from CCTV. Deep learning models are designed to predict the arrival time of pedestrians and vehicles, which is essential to calculate P-PET. Subsequently, the evaluation of the risk of pedestrians is carried out using specialized criteria tailored to different pedestrian categories based on P-PET values.

The contributions of this paper can be summarized as follows.

\begin{itemize}
    \item We propose a framework with computer vision technologies and predictive models to evaluate the pedestrian's potential risk in real time.
    \\
    \item We introduce a novel predicted surrogate safety measure, Predicted Post-Encroachment Time (P-PET), to accurately and interpretably evaluate the potential risk of pedestrians.
    \\
    \item We classified pedestrians into kids, adults, and cyclists to tailor specific evaluation rules to each pedestrian category, thus improving the performance of risk evaluation.
\end{itemize}

\subsection{Paper structure}
This paper is structured as follows. Section \ref{sec:section 2} outlines the methodology of our study, which is divided into four parts. Section \ref{sec:section 2.1} details the proposed framework for evaluating potential risks. Section \ref{sec:section 2.2} explains the preprocessing methods applied to the video dataset. Section \ref{sec:section 2.3} covers the preliminary definitions and introduces the deep learning models used for future state predictions. Section \ref{sec:section 2.4} defines P-PET and describes the algorithm that leverages P-PET for risk evaluation. Section \ref{sec:section 3} discusses the results from applying the proposed framework, with Section \ref{sec:section 3.1} examining the performance of the detection algorithm and the time distribution of pedestrian crossings by categories, Section \ref{sec:section 3.2} covering the training data and the performance of the arrival time prediction models, and Section \ref{sec:section 3.3} presenting the results of the risk evaluation algorithm using our proposed metrics. Finally, in Section \ref{sec:section 4}, conclusions and future work are presented.

\section{Methodology}
\label{sec:section 2}
\subsection{Framework}
\label{sec:section 2.1}
\begin{figure*}[!t]
  \centering
    \includegraphics[width=0.8\textwidth]{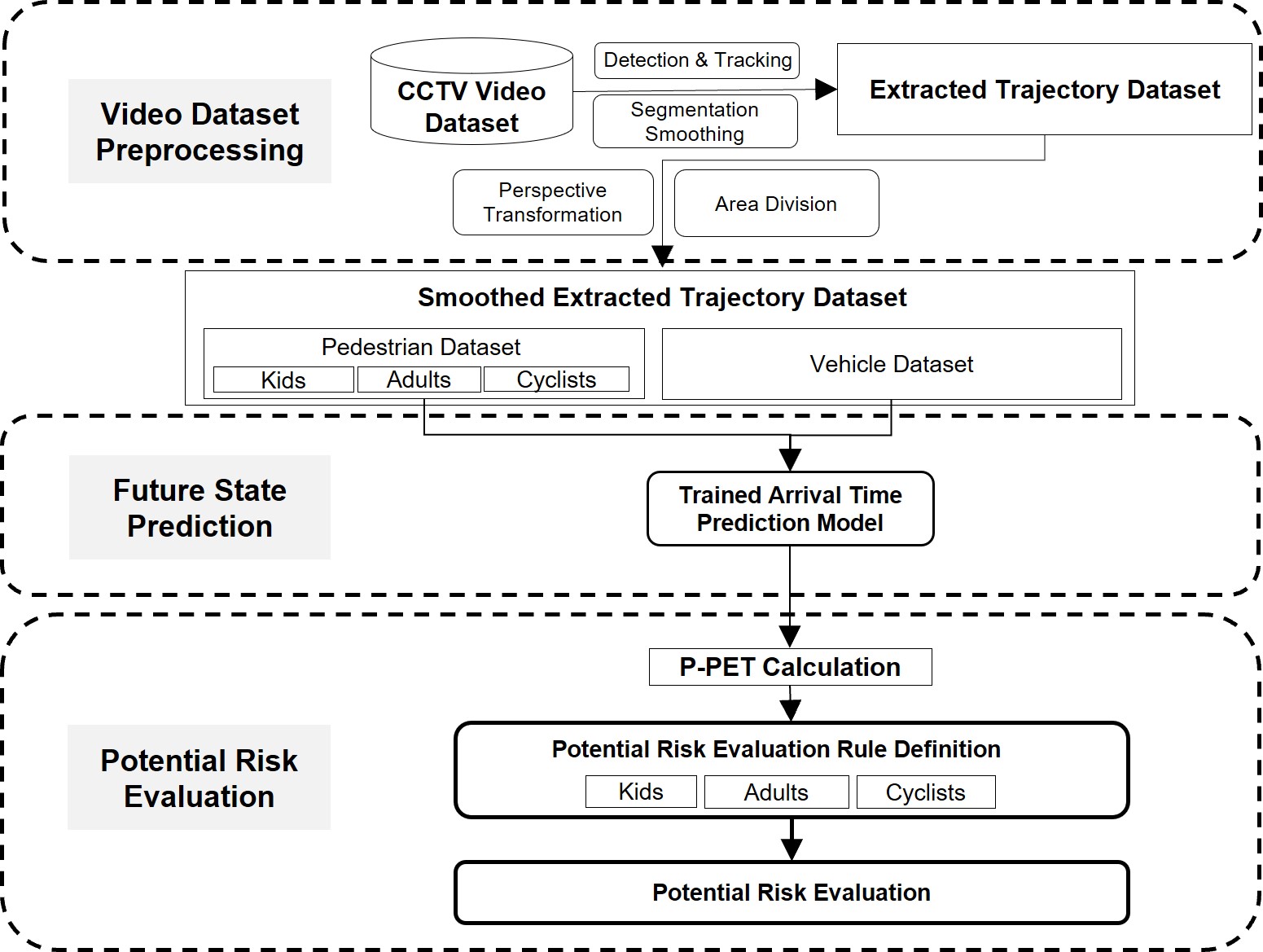}    
  \caption{Framework for Potential Risk Evaluation}\label{fig:Framework}
\end{figure*}

The proposed framework for the evaluation of pedestrian's potential risks is illustrated in Figure \ref{fig:Framework}. It comprises three key components: (i) Video Dataset Preprocessing (ii) Future State Prediction, and (iii) Potential Risk Evaluation. 
In \textit{Video Dataset Preprocessing}, we automatically convert CCTV video into the smoothed extracted trajectory dataset for vehicles and various categories of pedestrians.
In \textit{Future State Prediction}, this study uses trained deep learning models to predict the arrival time of pedestrians and vehicles to reach specific locations, critical to obtain the Predicted Post-Encroachment Time (P-PET).
In \textit{Potential Risk Evaluation}, we develop an algorithm with tailor-made specific evaluation rules based on P-PET to evaluate the potential risk of pedestrians.

\subsection{Video Dataset Preprocessing}
\label{sec:section 2.2}
% The study is carried at Sejong Non-signalized Intersection. A series of video data from CCTV at rush hours are provided.
\subsubsection{Site Introduction}
\begin{figure}[!ht]
  \centering
  \begin{subfigure}[b]{0.35\textwidth}
    \includegraphics[width=\textwidth]{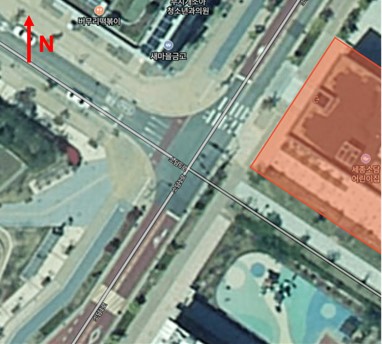}
    \caption{Bird's Eye View of the Target Intersection}
    \label{fig:bird-eye-view}
  \end{subfigure}
  \hfill % This adds some separation between the two subfigures
  \begin{subfigure}[b]{0.56\textwidth}
    \includegraphics[width=\textwidth]{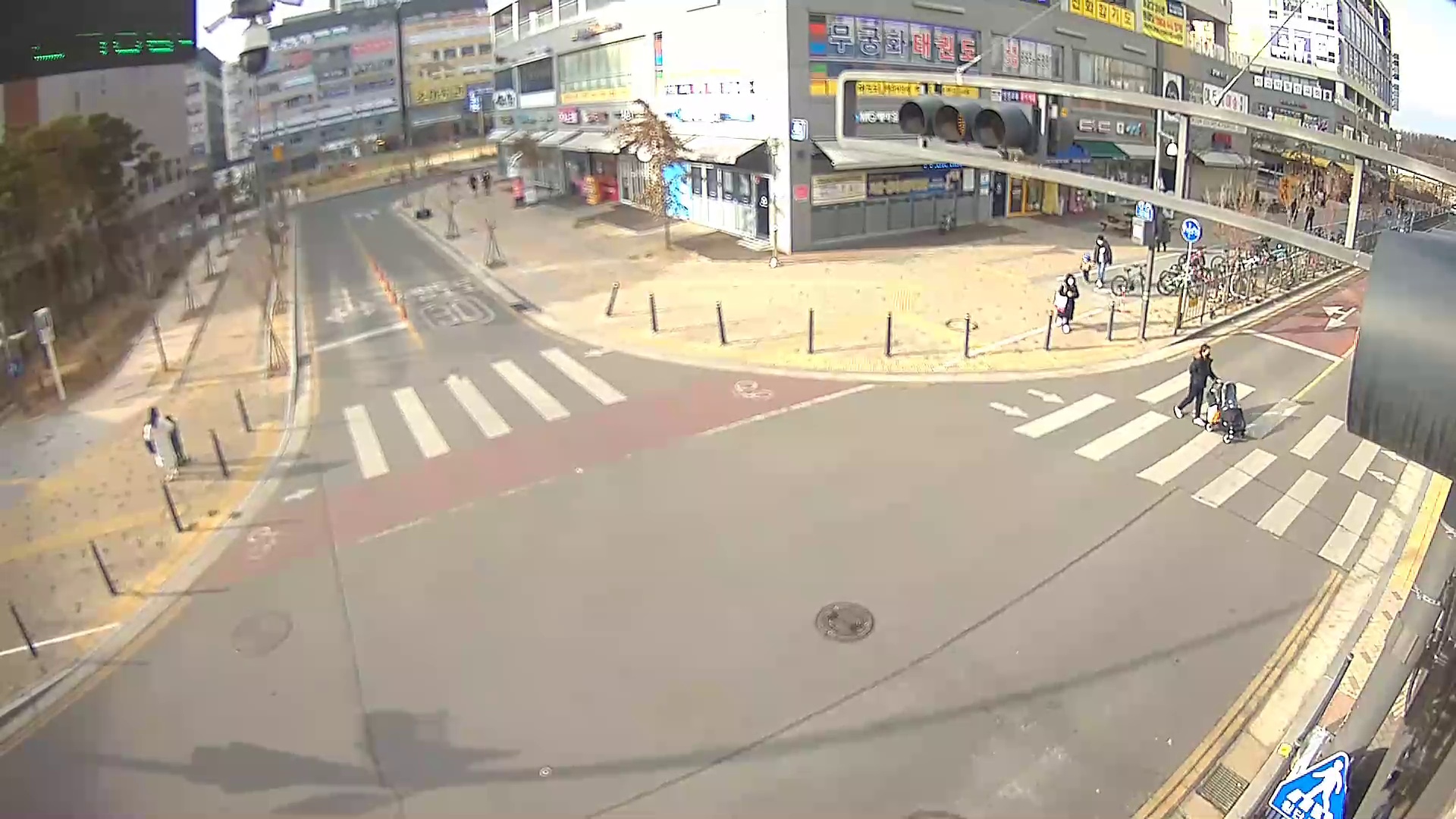}
    \caption{CCTV Camera View of the Target Intersection}
    \label{fig:cctv-view}
  \end{subfigure}
  \caption{Two views of the Target Intersection in Sejong City. (a) Bird's Eye View. (b) CCTV Camera View.}
  \label{fig:comparative-views}
\end{figure}

The intersection under this study is a T-shaped, non-signalized intersection located near a kindergarten in Sodam Dong, Sejong City, South Korea, as depicted in Figure \ref{fig:bird-eye-view}. 
The data for the study are extracted from a camera mounted at the intersection. The CCTV video dataset is collected from January $4^{th}$ to February $16^{th}$, from 4 to 6 pm. Each five-minute video file has a resolution of 1080 x 1920, containing 9000 frames at a frame rate of 30 FPS. 
In this study, we focus on three categories of pedestrians observed at the site: kids, cyclists, and adults. Adults are the majority of pedestrians crossing the intersection. Kids are deemed particularly vulnerable due to their unpredictable movement patterns and increased risk during crossing activities, requiring specific criteria sensitive to their safety. 
Cyclists, with their unique dynamics and behaviors different from those of kids and adults, also require distinct evaluations. These highlight the importance of creating evaluation rules specifically tailored to each category of pedestrians.

As depicted in Figure \ref{fig:cctv-view}, this study focuses on the Southwest-to-Northeast crossing due to the angle of the camera. The crosswalk spans approximately 11 meters, making it a relatively short non-signalized intersection. For pedestrians, the main concern at this intersection is the potential risk of collision with approaching vehicles.

\subsubsection{Video Processing with Computer Vision}

% 2023update: need to update the model with yolo v7 and also we label the dataset to increase the detection accuracy. And we dont have the perspective transformation for this job

In our research, we utilized the detection model YOLOv7 \citep{wang2023yolov7} to accurately identify objects and extract trajectory data of the adults, kids, cyclists and vehicles from video datasets. 
% This version in the YOLO series notably escalates detection accuracy without inflating the computational cost for real-time inferences, setting it apart from its predecessors\cite{wang2023yolov7}. 
To improve the performance of object detection, we fine-tuned the model. Leveraging Microsoft Azure \citep{Azure}, we annotated 1000 random frames of the video, tagging vehicles, adults, kids, and cyclists for model fine-tuning.

After object detection, object tracking is an essential algorithm that connects the information across successive frames for extracting pedestrian trajectories. For this purpose, we utilized Deep Simple Online and Real-time Tracking (Deep SORT) as our object tracking algorithm \citep{wojke2017simple}. Deep SORT leverages the advantage of both the Kalman Filter to consider the motion feature of the pedestrian and Matching Cascade to consider the pedestrian exterior appearance feature. 
    
\begin{figure}[!ht]
  \centering
  % First subfigure
  \begin{subfigure}[b]{0.13\textwidth}
    \includegraphics[width=\textwidth]{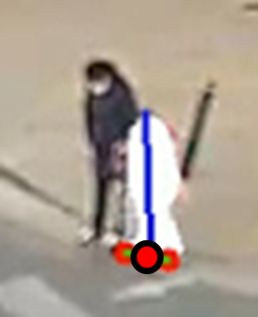}
    \caption{Pedestrian}
    \label{fig:Pedestrian}
  \end{subfigure}
  \hspace{3mm} % This will ensure that the remaining space is evenly distributed between the subfigures
  % Second subfigure
  \begin{subfigure}[b]{0.268\textwidth}
    \includegraphics[width=\textwidth]{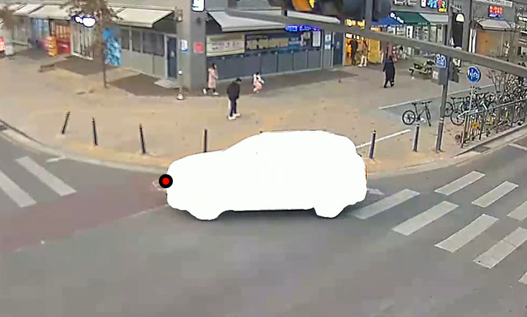}
    \caption{Vehicle (Right)}
    \label{fig:Vehicle-Right-to-Left}
  \end{subfigure}
  \hspace{3mm} % Even space
  % Third subfigure
  \begin{subfigure}[b]{0.13\textwidth}
    \includegraphics[width=\textwidth]{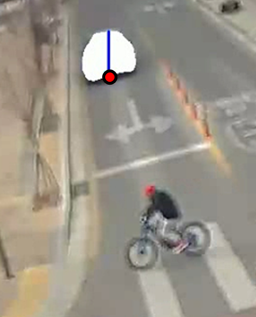}
    \caption{Vehicle (Up)}
    \label{fig:Vehicle-Up-to-Down}
  \end{subfigure}
  \hspace{6mm} % Even space
  % Fourth subfigure
  \begin{subfigure}[b]{0.268\textwidth}
    \includegraphics[width=\textwidth]{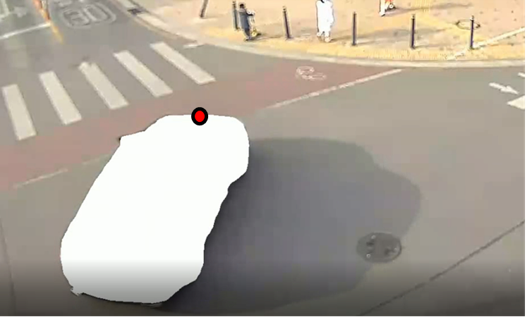}
    \caption{Vehicle (Left)}
    \label{fig:Vehicle-Left-to-Right}
  \end{subfigure}

  \caption{Segmentation Target Points for Pedestrians and Vehicles in Different Directions.}
  \label{fig:Segmentation-Targets}
\end{figure}

% bounded box
Due to inaccuracies and fluctuations that arise from the use of bounding boxes to define pedestrian locations, we use Segment Anything \citep{kirillov2023segany} for segmentation, ensuring a more precise location of pedestrians and vehicles beyond the limitations of bounding boxes. Since the collision point on the vehicle changes depending on the direction from which the vehicle approaches, we establish site-specific methodologies to automatically identify the locations of both vehicles and pedestrians within the segmentation results. For pedestrians in Figure \ref{fig:Pedestrian}, their position is marked by the midpoint between their feet. This is determined by finding the top edge and centroid of the segmentation area, which divides the pedestrian's image into right and left sides. Then, the lowest points on the left and right sides indicate the left and right feet, respectively. The pedestrian's location is the midpoint between these two points. For vehicles moving from right to left in Figure \ref{fig:Vehicle-Right-to-Left}, their position is marked by the leftmost point of the segmentation. For vehicles moving from top to bottom in Figure \ref{fig:Vehicle-Up-to-Down}, we first determine the centroid. The vehicle's position is identified as the point with the same x-coordinate as the centroid but at the lowest position. For vehicles moving from left to right in Figure \ref{fig:Vehicle-Left-to-Right}, their position is marked by the highest point.

\begin{figure}[!ht]
  \centering
  % First row of images
  \begin{subfigure}[b]{0.22\textwidth}
    \includegraphics[width=\textwidth]{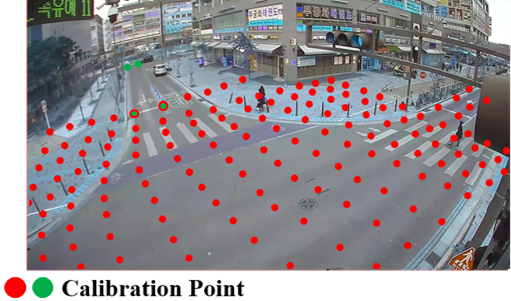}
    \caption{Calibration Points on the Image}
    \label{fig:manual-calibration-points}
  \end{subfigure}
  \hspace{3mm} % Adjust this value as needed to control the space between the first and second image
  \begin{subfigure}[b]{0.21\textwidth}
    \includegraphics[width=\textwidth]{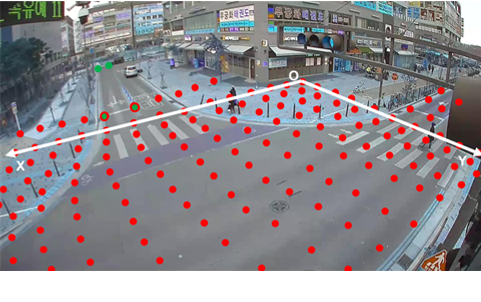}
    \caption{World Coordinate on the Image}
    \label{fig:world-coordinate-image}
  \end{subfigure}
  \hspace{3mm} 
  \begin{subfigure}[b]{0.22\textwidth}
    \includegraphics[width=\textwidth]{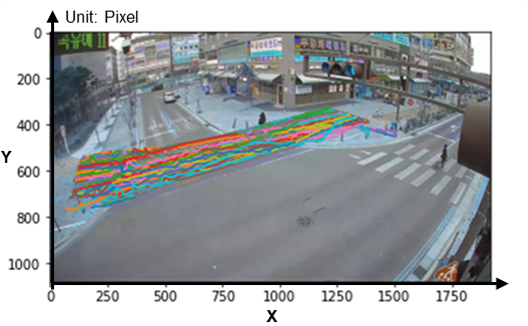}
    \caption{Trajectories Example on the Pixel Coordinate}
    \label{fig:trajectories-pixel-coordinate}
  \end{subfigure}
  \hspace{3mm} % Adjust this value as needed for the second row
  \begin{subfigure}[b]{0.22\textwidth}
    \includegraphics[width=\textwidth]{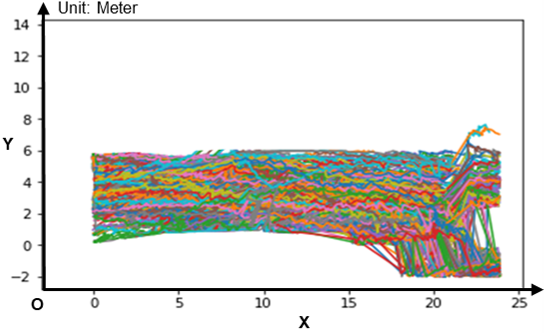}
    \caption{Trajectories Example on the World Coordinate}
    \label{fig:trajectories-world-coordinate}
  \end{subfigure}

  \caption{Calibration figures that demonstrate calibration points, world coordinates, and trajectory examples in both pixel and world coordinates.}
  \label{fig:calibration-figures}
\end{figure}

% Perspective Transfer 
Perspective transformation is conducted to create a projection from pixel coordinate to world coordinate. The captured pixel coordinate information from the video must be converted to a world coordinate to obtain the exact location in the real world. Therefore, we propose a perspective transformation method called ''Transformation Matrix'' ($M_T$) \citep{bradski2008learning, noh2022novel} in the OpenCV library that can be derived from four pairs of corresponding ''anchor points'' in the pixel coordinates (oblique view) and virtual world coordinates (overhead view).  
148 red points and 4 green points are calibrated as shown in Figure \ref{fig:manual-calibration-points}. Each four red points represent a 2x2-meter square in real life. Four green points present another 46x2-meter rectangular in real life. With 121 rectangular areas, the OpenCV function \textit{ getPerspectiveTransform()} is used. Therefore, 121 values of $M_T$ are obtained for each square to transfer the trajectory information from the coordinate of the pixels, shown in Figure \ref{fig:trajectories-pixel-coordinate} (c), to the coordinate of the world, shown in Figure \ref{fig:trajectories-world-coordinate} (d).

\subsubsection{Additional Information for Behavior Analysis and Evaluation}
\label{sec:section3.1.3}
Through the application of object detection, tracking, segmentation, and perspective transformation, we automate the extraction of trajectory data for vehicles, adults, kids, and cyclists. Despite the efficacy of automation, these video clips contain a wealth of additional information that is beneficial to assess the pedestrian's potential risk, which automated methods alone cannot fully exploit.

In complement to our automated processes, we delve deeper into the dataset through a meticulous observation of each pedestrian trajectory's video clips. This observation allows us to capture details of pedestrian behavior and interactions that automated techniques might overlook. Our detailed annotation of each trajectory, as outlined in Table \ref{tab:my-table}, enriches our dataset. These enriched data serve multiple purposes: they aid in the evaluation of our arrival time prediction models and risk evaluation algorithm, and facilitate the analysis of behavioral differences among pedestrians. Importantly, while this information is crucial for our comprehensive evaluation and analysis, it is not required for the automated processing component.

In this research work, we focus on the influence of the following variables.
\begin{itemize}
    \item \textbf{Sub-Classification}:
To assess the effectiveness of our pre-trained YoloV7 models in pedestrian classification and detection, we conducted observations of pedestrian subclasses. We labeled pedestrians to establish ground-truth data to evaluate detection model performance and for further analysis of different pedestrian types. All subsequent analyses of the impact of various types of pedestrians rely on these ground truth data, rather than on the results detected by the models, to eliminate errors that could arise from incorrect detection outcomes.
    \item \textbf{Pedestrian Awareness}:
We check whether a pedestrian is aware of an approaching vehicle by observing specific behaviors indicative of their awareness. Our criteria for determining whether a pedestrian has noticed an approaching vehicle include visible eye contact or head rotation towards the vehicle. If a pedestrian clearly directs their eyesight towards the vehicle or exhibits a significant head turn to check their surroundings, we record that they have acknowledged the vehicle's presence. Furthermore, if a vehicle enters a pedestrian's field of vision and the pedestrian significantly changes their motion in response, we also consider this as an indication of the pedestrian noticing the vehicle. The reason for this labeling is to separate pedestrians who have noticed approaching vehicles from our arrival time prediction model training dataset. The trained models aim to predict the pedestrian's future state assuming that they do not notice the approaching vehicle.  Therefore, in this framework, we evaluate the potential risk of the pedestrian under the assumption that the target pedestrian is not aware of the approaching vehicle, recognizing that these pedestrians are at a higher risk than those who are aware, even if they might actually be aware.

    \item \textbf{Time for Pedestrian Awareness}:
These data give us the precise moment when pedestrians become aware of an approaching vehicle, helping us to obtain the common location where pedestrians sense potential danger and typically recognize approaching vehicles in real scenarios. This collected information is utilized for the task of \textbf{Area Division}, which will be discussed later. 

    \item \textbf{Risk Level}:
To determine the optimal threshold of the Predicted Post-Encroachment Time and evaluate the performance of the potential risk evaluation algorithm, we classify the risk level of each pedestrian based on the rules of the vehicle-to-pedestrian conflict \citep{govinda2022pedestrian}. 

(1) Risk Level 0 (No conflict): No vehicle is moving nearby when the pedestrian crosses the intersection. The possibility of an accident is 0.

(2) Risk Level 1 (Normal conflict): Both pedestrians and vehicles move at their normal speeds. The possibility of an accident is very low in this situation.

(3) Risk Level 2 (Severe conflict):
In this situation, either one or both road users must stop or change their movements. This could mean slowing down, speeding up, changing direction, or even waving at the vehicle to avoid an accident. The chances of an accident happening here are pretty high.

\end{itemize}
%
% ###########################################################################################
% Please add the following required packages to your document preamble:
% \usepackage{graphicx}
\begin{table}[]
\caption{Additional Information}
\label{tab:my-table}
\centering
\resizebox{0.95\textwidth}{!}{%
\begin{tabular}{lll}
\hline
Variables            & Description                                                                                                                                                                                                                                                                                                                                                                                                                                                                                                                               & Catagory                                                                                                                                                    \\ \hline
ID                   & The id of the pedestrian after video processing                                                                                                                                                                                                                                                                                                                                                                                                                                                                                                                                                                                             &Integral  
               \\ \cline{2-3} 
Sub-Classification      & Pedestrian's Classification                                                                                                                                                                                                                                                                                                                                                                                                                                                                                                               & \begin{tabular}[c]{@{}l@{}}1. Cyclist\\ 2. Kid\\ 3. Adult\end{tabular}                                                                                           \\ \cline{2-3} 
Pedestrian Awareness           & Whether the pedestrian notice the coming vehicle or not                                                                                                                                                                                                                                                                                                                                                                                                                                                                                              & \begin{tabular}[c]{@{}l@{}}0. No or no vehicle\\ 1. Pedestrian notice the coming vehicle\end{tabular}                                                                                            \\ \cline{2-3} 
Time for Pedestrian Awareness      & The time when the pedestrian notice the coming vehicle
& Float                                          
\\ \cline{2-3} 
Reaction Type        & Pedestrian's reaction to the coming vehicle.                                                                                                                                                                                                                                                                                                                                                                                                                                                                                              & \begin{tabular}[c]{@{}l@{}}0. No Obvious Action\\ 1. Decelerate to Wait\\ 2. Accelerate to Pass\end{tabular}                                                        \\ \cline{2-3} 
 
Risk Level & \begin{tabular}[c]{@{}l@{}}Risk of the Pedestrian \end{tabular} & \begin{tabular}[c]{@{}l@{}}0. Risk 0\\ 1. Risk 1\\ 2. Risk 2\end{tabular}                                                                                        \\ \hline
                     &                                         
\end{tabular}%
}
\end{table} 

\subsubsection{Area Division}
\label{sec:section3.1.4}
\begin{figure}[!ht]
  \centering
  \includegraphics[width=0.7\textwidth]{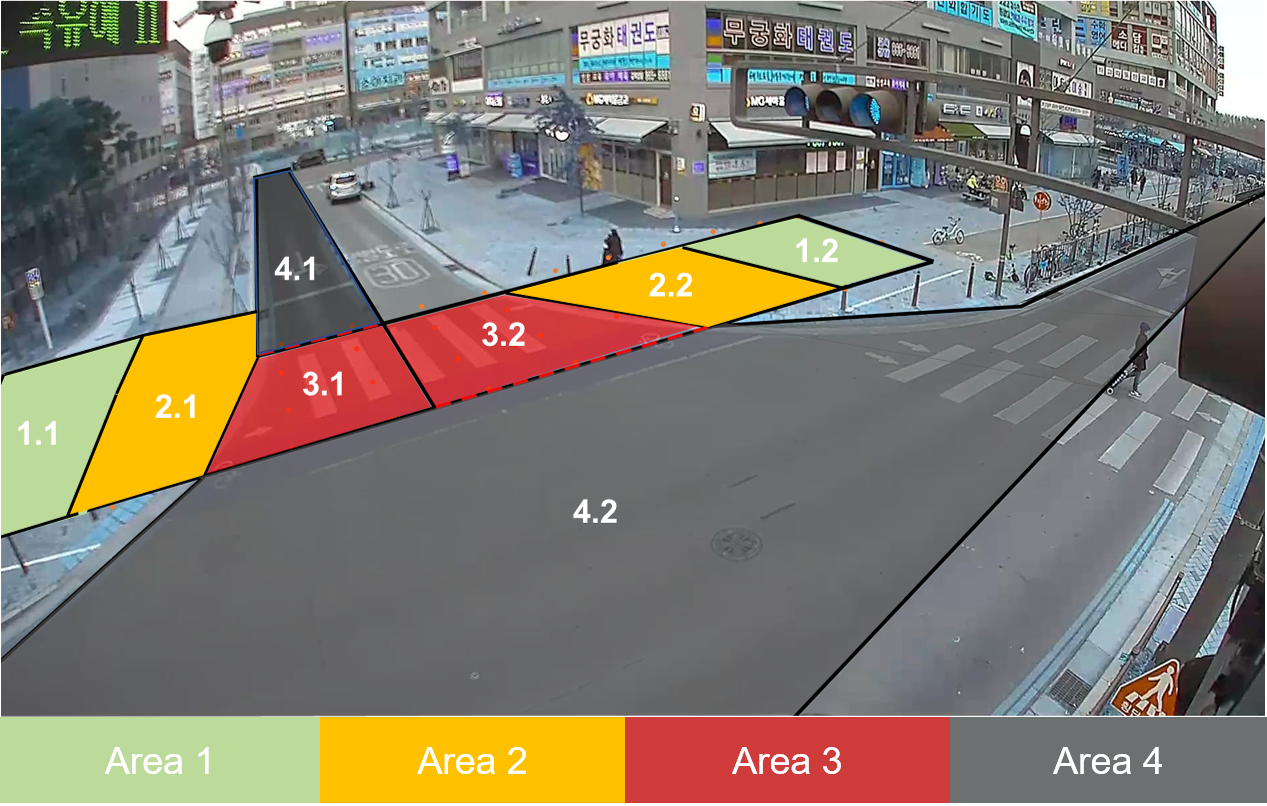}
  \caption{Illustration of Area Division Rules Applied to the Site}\label{fig:area-division}
\end{figure}

% Area division is updated by the attention graph
Designing a proactive protection system for non-signalized intersections presents a formidable challenge, largely due to the intricate behaviors of pedestrians. Inspired by \cite{jayaraman2020analysis}, which simplifies pedestrian crossing behaviors into four sequential steps: approaching, reacting, crossing, and exiting the crosswalk, we adapt this conceptual framework to our context. Although our division of the intersection into four areas, depicted in Figure \ref{fig:area-division}, draws inspiration from these steps, it does not replicate them exactly. Instead, this division guides the development of our risk evaluation algorithm, ensuring that it aligns with the dynamics of pedestrian crossing behavior.

\begin{itemize}
    \item \textbf{Area 1 (Pedestrian approach to the crosswalk)}: 
    Area 1 is defined as the zone where pedestrians approach the intersection, at which stage their behavior remains uninfluenced by coming vehicles. To evaluate the potential risk to the target pedestrian in this area, we collect the trajectory data for both the pedestrian heading toward the intersection and the nearby vehicles. These data are crucial to predict the arrival time. Unlike Area 2, the risk evaluation algorithm is specifically designed not to assess risk in Area 1. This decision aims to prevent the algorithm from becoming overly sensitive and to minimize computational costs.
    
    % The pedestrian in Area 1 didn`t get influence from either the coming vehicle or their own traffic concerns.This walking state of non-influence by the coming vehicle and safety concerns is consider as Normal walking behavior. In Area 1, all pedestrian is considered as in Normal Walking state 
    
    \item \textbf{Area 2 (Pedestrian react to approaching vehicles)}: 
    Area 2, positioned near the intersection, represents a critical phase where pedestrians are on the verge of crossing but have not yet begun. In this area, they are aware of approaching vehicles and may alter their behavior accordingly to ensure a safe crossing. Their reactions might range from accelerating to cross more rapidly, maintaining their pace as if still in Area 1, or decelerating to await a safer opportunity to cross. Highlighting the potential risk to pedestrians in this zone becomes essential, especially for those who fail to recognize an approaching vehicle or choose riskier behaviors, such as speeding up or continuing at their current pace in high-risk scenarios. The demarcation between Areas 1 and 2 is guided by the intersection's geometry and the observed location at which pedestrians typically become alert to approaching vehicles.

    %According to that, the pedestrians in Area 2 are the target pedestrian that the warning system performed to influence their decision to improve the pedestrian safety.
    
    \item \textbf{Area 3 (Pedestrian cross the intersection)}: 
Area 3 is designated as the active crossing zone of the intersection, identified as the segment with the highest risk of vehicle-pedestrian collisions and is referred to as the \textbf{"Conflict Area"}. This area spans the vehicle's path in two directions, resulting in two principal conflict areas where pedestrian and vehicle paths intersect. In this study, the initial conflict area entered by a pedestrian is termed the \textbf{"Closer Area"}, while the subsequent one is termed the \textbf{"Further Area"}. 
For example, a pedestrian traversing from left to right across the intersection would encounter areas in the sequence of 1.1, 2.1, 3.1 (Closer Area), and then 3.2 (Further Area). Recognizing that the challenge of making predictions escalates with distance, our research explores the differences between these two areas, aiming to understand their impact on prediction accuracy and risk evaluation performance.
    
    \item \textbf{Area 4 (Vehicle approaching the conflict area)}: 
Area 4 acts as a precursor zone where vehicles approach the conflict area before entering it. Within our study's setting, vehicles display distinct movement patterns based on their direction. Specifically, Area 4.1 caters to vehicles moving from the northwest to the southeast (depicted as top to bottom), while Area 4.2 is tailored for vehicles transitioning from the northeast to the southwest or the reverse (illustrated as left-to-right and right-to-left). The predominantly straight trajectories in Area 4.1 contrast with the mainly curved paths in Area 4.2. This distinction is critical for training our predictive model, which enables it to forecast arrival time with greater precision by acknowledging differences in vehicle movement patterns.

\end{itemize}

\subsection{Future State Prediction}
\label{sec:section 2.3}
%link
% After Data Collection and Preprocessing, and Pedestrian Classification, we fully extract all the needed information. 
% Data Collection and Preprocessing, along with Pedestrian Classification, retrieve all the required information from the video dataset to evaluate the potential risk.
% %
% The proactive protection system must foresee the upcoming threats to protect the pedestrian before a collision occurs.
% %
% Among the definition of SSMs, which surve as one of the most widely used methods for identifying future threats \cite{tak2018comparison}, arrival time at a specific location plays an important role. 
% %
% Therefore, in this section, we develop a time-sequence prediction model to predict pedestrian arrival time to evaluate the pedestrians' potential risk.
Video Dataset Preprocessing  extract and process all necessary trajectory information from our video dataset, aimed at evaluating pedestrian's potential risk.
A potential risk evaluation system must be adept at predicting future states to identify imminent threats, thus enabling proactive measures to protect pedestrians before a crash occurs. Furthermore, arrival time at specific locations is essential in defining Surrogate Safety Measures (SSMs), which represents one of the most crucial variables to detect future conflicts\citep{tak2018comparison}.
Therefore, in this research, we employ deep learning approaches to predict the arrival time of both pedestrians and vehicles, contributing to the evaluation of pedestrian risk.
It is important to note that, as mentioned in the definition of \textbf{\emph{Pedestrian Awareness}}, the framework is designed with the assumption that target pedestrians are unaware of approaching vehicles. As such, deep learning models are trained on the dataset with pedestrians who have not noticed approaching vehicles.
\subsubsection{Preliminaries}
\label{sec: section3.2.1}
\begin{figure}[!ht]
  \centering
  % First row of images
  \begin{subfigure}[b]{0.4\textwidth}
    \includegraphics[width=\textwidth]{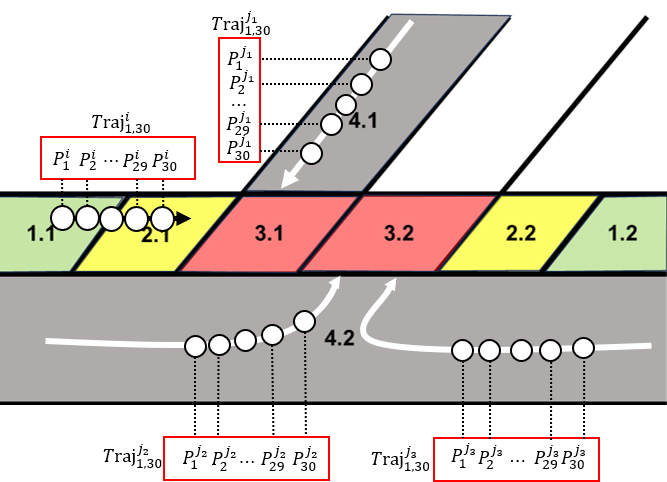}
    \caption{The Trajectory Input for Pedestrian Crossing from Left to Right}
    \label{fig:Input for Pedestrian Crossing from Left to Right }
  \end{subfigure}
  \hspace{5mm} % Space between the first and second image
  \begin{subfigure}[b]{0.4\textwidth}
    \includegraphics[width=\textwidth]{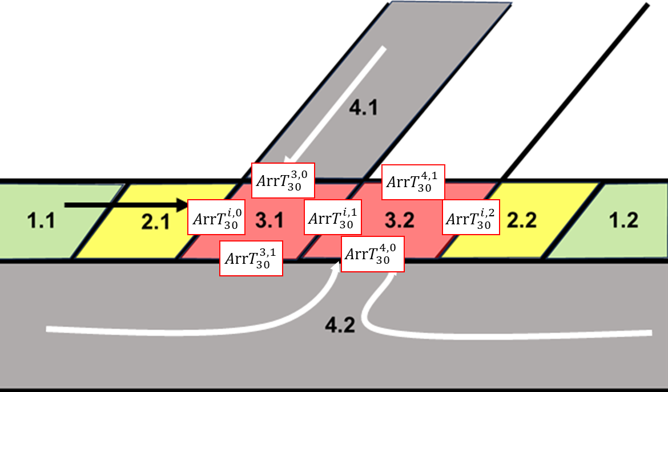}
    \caption{The Arrival Time Output for Pedestrian Crossing from Left to Right}
    \label{fig:Arrival Time Output for Pedestrian Crossing from Left to Right}
  \end{subfigure}
  
  % Add some vertical spacing between the rows
  \vspace{1cm}
  
  % Second row of images
  \begin{subfigure}[b]{0.4\textwidth}
    \includegraphics[width=\textwidth]{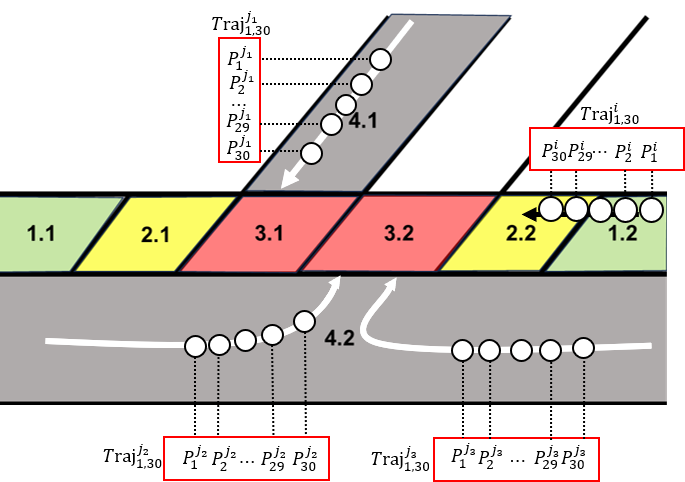}
    \caption{The Trajectory Input for Pedestrian Crossing from Right to Left}
    \label{fig:Trajectory Input for Pedestrian Crossing from Right to Left}
  \end{subfigure}
  \hspace{5mm} % Space between the third and fourth image
  \begin{subfigure}[b]{0.4\textwidth}
    \includegraphics[width=\textwidth]{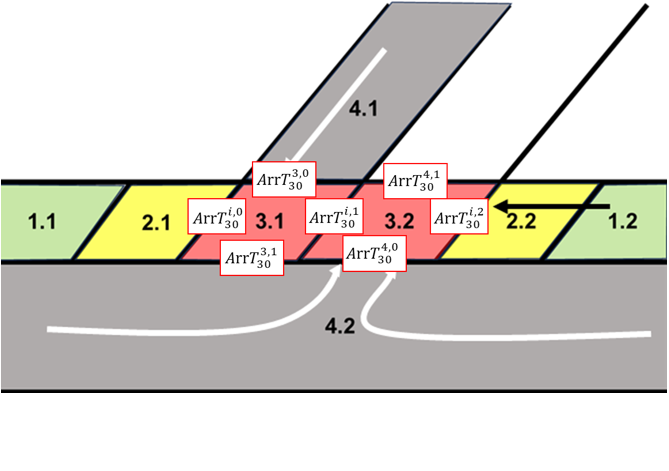}
    \caption{The Arrival Time Output for Pedestrian Crossing from Right to Left}
    \label{fig:Arrival Time Output for Pedestrian Crossing from Right to Left}
  \end{subfigure}

  \caption{Illustration of the Prediction Process: (a)(c) Trajectory Input.
(b)(d) Predicted Arrival Time Output.}
  \label{fig:Illustration of the Prediction Process}
\end{figure}

In this section, we introduce several key definitions used throughout our study:

\begin{itemize}
    \item\textbf{\emph{Definition 1}} (Target Agent): This article defines the target agent $i$ as an individual pedestrian or an individual vehicle, distinguished based on their dynamic behavior. Specifically, $i=0$ represents adults, $i=1$ denotes kids, $i=2$ corresponds to cyclists, $i=3$ refers to vehicles in Area 4.1, and $i=4$ identifies vehicles in Area 4.2.

    \item\textbf{\emph{Definition 2}} (Complete Trajectory): Consider a complete trajectory of a Target Agent $i$ extracted from the video defined as $\text{Traj}_N^i = \{P_1^i, \cdots, P_n^i, \cdots, P_N^i\}$, where $N$ is the length of the complete trajectory, $1 \leq n \leq N$. Each point $P_n^i = (x_n, y_n, t_n)$ has information in world coordinates and time step. 

    \item\textbf{\emph{Definition 3}} (Sliding Window Trajectory): In the context of this study, where the frame rate (FPS) of the video dataset is 30 frames per second, we define a sliding window trajectory for a pedestrian or vehicle $i$ as a sequence of consecutive points extracted from the complete trajectory $\text{Traj}_N^i$, using a fixed window size of $m = 30$ points. This window size corresponds to a temporal span of 1 second, reflecting the chosen FPS of the video. Thus, a sliding window trajectory can be denoted as $\text{Traj}_{j,m}^i = \{P_j^i, P_{j+1}^i, \cdots, P_{j+m-1}^i\}$, where $1 \leq j \leq N-m+1$. Each trajectory $\text{Traj}_{j,m}^i$ represents a 1-second-long segment of movement, starting from point $P_j^i$ and including $m$ consecutive points up to $P_{j+m-1}^i$.

    \item\textbf{\emph{Definition 4}} (Target Location): In this study, we define the target locations of pedestrians and vehicles with different identifiers $q$. For pedestrians, there are three target locations: $q=0$ indicates that the pedestrian enters the conflict area, $q=1$ denotes the pedestrian crossing the center line of the crossroad, and $q=2$ means that the pedestrian leaves the conflict area. For vehicles, two target locations are defined: $q=0$ means that the vehicle enters the conflict area and $q=1$ indicates that the vehicle leaves the conflict area.

    \item\textbf{\emph{Definition 5}} (Arrival Time): Let us consider the time it takes for a pedestrian or vehicle $i$ at the location $P_{j+m-1}^i$ to arrive at a specific target location $q$. This is represented as $\text{ArrT}_{j+m-1}^{i,q} = T_q^i - T_{P_{j+m-1}^i}^i$, where $T_q^i$ represents the time in which the pedestrian or vehicle $i$ reaches the target location $q$, and $T_{P_{j+m-1}^i}^i$ denotes the time corresponding to the pedestrian or vehicle $i$ being at location $P_{j+m-1}^i$. This formulation allows us to calculate the exact time it takes to reach a designated target location from the current position using complete trajectory information.

    \item\textbf{\emph{Definition 6}} (Labeled Dataset): A labeled data set $D_{\text{Labeled}}^{q}$ for the prediction of arrival time consists of pairs of Sliding Window Trajectories and their corresponding Arrival Time at a specific target location $q$, that is, $D_{\text{Label}}^{q} = \{(\text{Traj}_{j,m}^{i}, \text{ArrT}_{j+m-1}^{i,q})\}$. Each pair in the dataset includes a sliding window trajectory $\text{Traj}_{j, m}^{i}$, which is a sequence of consecutive points representing a portion of the entire trajectory of a pedestrian or vehicle $i$, and the arrival time $\text{ArrT}_{j+m-1}^{i,q}$, which indicates the time it takes for the entity at the end of this window to arrive at a specific target location $q$.

     \item\textbf{\emph{Definition 7}} (Arrival Time Prediction): As shown in Figure \ref{fig:Illustration of the Prediction Process}, the process $\text{Traj}_{j,m}^i \rightarrow \text{ArrT}_{j+m-1}^{i,q}$ predicts the time required for an agent $i$ to reach a specific target location $q$ from its position at $P_{j+m-1}^i$, based on a historical sliding window trajectory $\text{Traj}_{j,m}^i$. Essentially, the arrival time prediction transforms the input sliding window trajectory $\text{Traj}_{j,m}^i$ into the predicted arrival time $\hat{\text{ArrT}}_{j+m-1}^{i,q}$.

    \item\textbf{\emph{Definition 8}} (Arrival Time Prediction Models): We define an Arrival Time Prediction Model $\text{ATPM}_{q}^{i}$ that transforms the input sliding window trajectory $\text{Traj}_{j,m}^i$ into the predicted arrival time $\hat{\text{ArrT}}_{j+m-1}^{i,q}$. Formally, $\hat{\text{ArrT}}_{j+m-1}^{i,q} = \text{ATPM}_{q}^{i}(\text{Traj}_{j,m}^i)$, where $\hat{\text{ArrT}}_{j+m-1}^{i,q}$ denotes the model prediction for the time required for the target agent $i$ to reach the target location $q$ from its position at $P_{j+m-1}^i$.

\end{itemize}

\subsubsection{Arrival Time Prediction}

% Currently, many deep learning methods are promoted to solve the time series prediction since they can efficiently capture the spatial-temporal feature of big trajectory data. 
As various deep learning approaches easily capture the spatial-temporal features of large-scale trajectory data, they are currently advocated as a solution to the time series prediction problem.
This article utilizes and compares multiple deep learning models, including Gated Recurrent Unit (GRU), Long Short-Term Memory(LSTM), and Transformer, to accurately forecast the pedestrian's arrival time. Furthermore, to provide a comprehensive evaluation of these deep learning models, we introduce a comparative analysis with a baseline algorithm. This algorithm, called the Historical Average (HA), forecasts the arrival time based on the historical average velocity and direction of the pedestrian path.
\begin{itemize}

    \item\textbf{\emph{HA}}: 
    
    The Historical Average (HA) algorithm estimates the time it takes for the target agent to reach the conflict area, assuming a constant velocity. The algorithm is detailed as follows: For trajectory $\text{Traj}_{j,m}^i$, a direction vector $\vec{D}_i$ is computed by subtracting the start point from the end-point coordinates. Let $\vec{P}_{\text{start}}^i$ and $\vec{P}_{\text{end}}^i$ represent the starting and ending points of the trajectory $\text{Traj}_{j,m}^i$, respectively. The direction vector $\vec{D}_i$ and its norm $||\vec{D}_i||$ are given by:
    
    \begin{equation}
    \vec{D}_i = \vec{P}_{\text{end}}^i - \vec{P}_{\text{start}}^i
    \label{eq:direction_vector_ha}
    \end{equation}
    
    \begin{equation}
    ||\vec{D}_i|| = \sqrt{(P_{\text{end}, x}^i - P_{\text{start}, x}^i)^2 + (P_{\text{end}, y}^i - P_{\text{start}, y}^i)^2}
    \label{eq:norm_direction_vector_ha}
    \end{equation}
    
    The direction angle relative to the x-axis, $\theta_i$, is calculated using the equation:
    \begin{equation}
    \theta_i = \text{atan2}(\vec{D}_i[y], \vec{D}_i[x])
    \label{eq:direction_angle_ha}
    \end{equation}
    The average velocity ($V_{\text{avg}, i}$) for each trajectory is obtained by projecting the instantaneous velocities onto $\vec{D}_i$ and averaging them: 
    \begin{equation}
    V_{\text{avg}, i} = \frac{1}{n} \sum_{k=1}^{n} \vec{V}_k \cdot \frac{\vec{D}_i}{||\vec{D}_i||},
    \label{eq:avg_velocity_ha}
    \end{equation}
    where $\vec{V}_k$ represents the velocity at time step $k$, and $n$ is the total number of time steps.
    
    To estimate the arrival time to the specific target location ($\text{ArrT}_{j+m-1}^{i,q}$), the algorithm considers the boundary lines of the target location. Assuming that $d_i$ represents the shortest distance from the final point of the trajectory to the target line and $V_{\text{avg}, i} \cos(\theta_i - \phi)$ represents the component of the average velocity in the direction of the target line ($\phi$ being the angle of the line with respect to the x axis), the time to reach the line is estimated as:
    
    \begin{equation}
    \text{ArrT}_{j+m-1}^{i,q} = \frac{d_i}{V_{\text{avg}, i} \cos(\theta_i - \phi)}
    \label{eq:arrival_time_ha}
    \end{equation}

    \item\textbf{\emph{LSTM}}\citep{hochreiter1997long}:
    
    LSTMs are sophisticated variants of Recurrent Neural Networks specifically designed to capture long-term dependencies within sequential data. They feature an intricate gating mechanism, including Forget, Input, and Output gates, that regulates the flow of information, making them exceptionally well-suited for analyzing time series data such as pedestrian movement trajectories. Their ability to maintain historical information over extended periods makes them ideal for accurate time predictions.
    
    \item\textbf{\emph{GRU}}\citep{chung2014empirical}:
    
    GRUs streamline the recurrent network architecture by combining the cell state and hidden state into a unified structure and reducing the number of gates to just two: Update and Reset. This simplification not only decreases the complexity of the model, but also reduces the number of parameters, enhancing the processing speed without sacrificing the ability to discern and learn from temporal patterns. GRUs are particularly effective in scenarios where dataset size is constrained, but temporal accuracy is paramount.
    
    \item\textbf{\emph{Transformer}}\citep{vaswani2017attention}:
    
    The Transformer model abandons traditional recurrence mechanisms in favor of attention-based strategies, enabling it to directly model relationships between all parts of the sequence simultaneously. This approach allows for significant improvements in computation efficiency and model scalability. Notably adept at handling tasks that require a deep understanding of the entire context, such as predicting arrival times from complex trajectories, Transformers provide superior performance, particularly in settings with extensive data.

\end{itemize}

In this study, the labeled dataset $D_{\text{Labeled}}^{q}$ is partitioned into a training dataset and a validation dataset using an 8:2 ratio. The selection of the deep learning architecture, whether LSTM, GRU, or Transformer, is determined by evaluating their performance based on the Mean Absolute Error (MAE) loss during the validation process. This approach ensures that the optimal model is chosen for the Arrival Time Prediction Model $\text{ATPM}_{q}^{i}$. Therefore, we formalize the selection criterion as follows:

\begin{equation}
\label{eqn:Optimum Arrival Time Two Lines}
\begin{split}
&\text{ATPM}{q}^{i} = \underset{\text{model} \in \{\text{LSTM}, \text{GRU}, \text{Transformer}\}}{\mathrm{arg\,min}} \ \text{MAE}_{\text{validation}}(\text{model}) \\
\end{split}
\end{equation}

This methodology allows us to identify the most accurate model for predicting arrival time. Once the best model is selected, the predicted arrival time, $\hat{\text{ArrT}}_{j+m-1}^{i,q}$, for any given sliding window trajectory $\text{Traj}_{j,m}^i$ is calculated as:

\begin{equation}
\hat{\text{ArrT}}_{j+m-1}^{i,q} = \text{ATPM}_{q}^{i}(\text{Traj}_{j,m}^i),
\end{equation}

where $\hat{\text{ArrT}}_{j+m-1}^{i,q}$ represents the model output, indicative of the time required for the target agent $i$ to reach the specific location $q$.

\subsection{Potential Risk Evaluation} 
\label{sec:section 2.4}
\subsubsection{Predicted Post-Encroachment Time Definition}
The use of Surrogate Safety Measures (SSMs) to estimate pedestrian and vehicle conflicts at intersections is increasingly popular thanks to their efficacy in reducing complex scenarios to simpler indicators. 
Researchers commonly employ Time-To-Collision (TTC) and Post-Encroachment Time (PET) as key SSMs when evaluating the safety of non-signalized intersections \citep{bonela2022review}. While TTC has a notable strength in objectively and quantitatively measuring the severity of the interaction \citep{chen2017surrogate}, its original design was intended to estimate the collision time between vehicles moving at their current speed in rear-end conflicts with overlapping trajectories. 
Post-Encroachment Time (PET), one of the most widely used SSMs, is often used to analyze pedestrian-vehicle interaction \citep{govinda2022pedestrian}. For example, \cite{fu2016pedestrian} classifies vehicle-pedestrian interactions with PET in less than 5 seconds as conflicts and in less than 1.2 seconds as dangerous conflicts.
Furthermore, \cite{marisamynathan2020pedestrian} sets PET threshold values to classify the intensity of the conflict between a pedestrian and a vehicle.
However, previous studies used PET values to classify conflict events and evaluate pedestrian safety based on past data for safety analysis rather than proactive protection of pedestrians \citep{he2018assessing, vedagiri2015traffic, zhang2017safety,ni2016evaluation,fu2016pedestrian}. Furthermore, traditional PET also shows shortcomings, such as its insufficient ability to accurately measure the severity of the conflict \citep{zhang2012pedestrian}.
Therefore, we propose a novel adjusted SSM, Predicted Post-encroachment Time (P-PET), that can be utilized to predict the severity of future pedestrian-vehicle conflicts and evaluate pedestrian's potential risk to safeguard pedestrians proactively.

Equation \ref{eqn:PET} shows the definition of traditional PET.

\begin{equation}
\label{eqn:PET}
\begin{split}
& {PET}_{PF} = T_{V,enter} - T_{P,leave}   \\
& {PET}_{VF} = T_{P,enter} - T_{V,leave}    \\
\end{split}
\end{equation}
where ${PET}_{PF}$ represents PET in Pedestrian-First Case, ${PET}_{VF}$ represents PET in Vehicle-First Case, $T_{P,enter}$ represents the time when the pedestrian enters the conflict area, $T_{P,leave}$ represents the time when the pedestrian leaves the conflict area, $T_{V,enter}$ represents the time when the vehicle enters the conflict area and $T_{V,leave}$ represents the time when the vehicle leaves the conflict area.

\begin{figure}[!ht]
\centering
\includegraphics[width=0.6\textwidth]{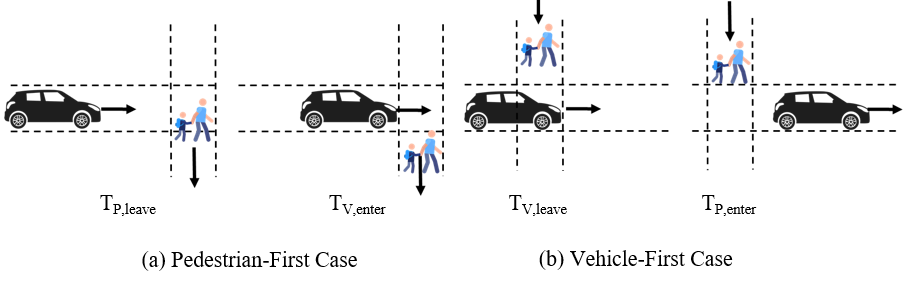}
\caption{Illustration of Post-Encroachment Time (PET) Definitions for Two Scenarios: Pedestrian First and Vehicle First.}\label{fig: PET_Details}
\end{figure}

Compared to PET, which is calculated from historical happened events, P-PET utilizes the predicted arrival time to predict the future state before the conflict happens. Therefore, P-PET can predict the level of conflict between the pedestrian and vehicle.  The way to calculate P-PET is defined by the following Equation \ref{eqn:P-PET}

\begin{equation}
\label{eqn:P-PET}
\begin{split}
& \textrm{P-PET}_{PF} = \hat{T}_{P,leave} - \hat{T}_{V,enter}    \\
& \textrm{P-PET}_{VF} = \hat{T}_{V,leave} - \hat{T}_{P,enter}    \\
\end{split}
\end{equation}
where  $\hat{T}_{P,enter} $, $\hat{T}_{P,leave} $ represent the predicted time when the pedestrian enters or leaves the conflict area based on the deep learning model, and $\Bar{T}_{V,enter}$, $\Bar{T}_{V,leave}$ represent the predicted time when the vehicle enters or leaves the conflict area based on the arrival time prediction model.

In coordination with Figure \ref{fig:Arrival Time Output for Pedestrian Crossing from Right to Left}, we refine the Predicted Post-Encroachment Time (P-PET) formulation to reflect the dynamics specific to different conflict areas and types of conflict encountered. P-PET is a critical metric for assessing potential collision scenarios between pedestrians and vehicles within designated conflict areas. The formulation for calculating P-PET, when a pedestrian traverses from left to right across the conflict area, is delineated as follows:

\begin{equation}
\label{eqn:P-PET_combine}
\begin{aligned}
\textrm{P-PET}_{C_{PF}} &= \hat{\text{ArrT}}_{j+m-1}^{3,0} - \hat{\text{ArrT}}_{j+m-1}^{i,1}, \\
\textrm{P-PET}_{C_{VF}} &= \hat{\text{ArrT}}_{j+m-1}^{i,0} - \hat{\text{ArrT}}_{j+m-1}^{3,1}, \\
\textrm{P-PET}_{F_{PF}} &= \hat{\text{ArrT}}_{j+m-1}^{4,0} - \hat{\text{ArrT}}_{j+m-1}^{i,2}, \\
\textrm{P-PET}_{F_{VF}} &= \hat{\text{ArrT}}_{j+m-1}^{i,1} - \hat{\text{ArrT}}_{j+m-1}^{4,1},
\end{aligned}
\end{equation}

where:
\begin{itemize}
    \item $\hat{\text{ArrT}}_{j+m-1}^{3,0}$ and $\hat{\text{ArrT}}_{j+m-1}^{3,1}$ signify the predicted time for a vehicle to enter and exit the conflict area 3.1, respectively.
    \item $\hat{\text{ArrT}}_{j+m-1}^{4,0}$ and $\hat{\text{ArrT}}_{j+m-1}^{4,1}$ denote the predicted time for a vehicle to enter and exit the conflict area 3.2, respectively.
    \item $\hat{\text{ArrT}}_{j+m-1}^{i,0}$ and $\hat{\text{ArrT}}_{j+m-1}^{i,2}$ represent the time predicted for a pedestrian of category i to enter the proximal and exit distal conflict areas, respectively.
    \item $\hat{\text{ArrT}}_{j+m-1}^{i,1}$ is the predicted time for a pedestrian to transition from the closer to the further conflict area.
\end{itemize}

Equation \ref{eqn:P-PET_combine} describes a method for calculating four specific P-PET (Predicted Post-Encroachment Time) values for each pedestrian as they cross two conflict areas, whether moving from left to right or vice versa. In P-PET analysis, a small negative value of ${P-PET}_{PF}$ signals a critical conflict scenario, indicating the possibility of a collision in which the vehicle could strike the pedestrian from the front. Similarly, the values of ${P-PET}_{CF}$, regardless of whether they are negative or positive, denote a significant conflict potential. Negative values warn of a frontal collision risk, whereas positive values reveal a pedestrian's approach to an already occupied crosswalk, underscoring the existence of a risk scenario.

\subsubsection{Potential Risk Evaluation Algorithm}
\label{sec:section3.3.2}
%% begining of the paragraph and connect with PET to the designed algorithm.
In this section, we detail an algorithm shown in pseudocode \ref{code:Risk}, aimed at pedestrian's potential risk evaluation at non-signal intersection in real time through Predicted Post-Encroachment Time (P-PET). This automated process, from initial video data processing to determination of potential pedestrian risk, is designed based on the study of pedestrian crossing behaviors, illustrated in Figure \ref{fig:area-division}. With the ability to continuously process video data, the algorithm supports the ongoing generation of results, ensuring that it can effectively offer proactive real-time protection to pedestrians. 

 Continuous video footage is used as the initial input, applying a suite of tools, detection, tracking, segmentation, perspective transformation, and area division, to selectively extract and store the movement data of pedestrians and vehicles in each frame. It focuses specifically on pedestrians and vehicles that enter a designated area of interest, as shown in Figure \ref{fig:area-division} and referred to as the Target Agent, following the criteria of Definition 1 in Section \ref{sec: section3.2.1}. This selective approach not only refines our analysis target on pedestrians who intend to cross, thereby enhancing the relevance to pedestrian crossing safety, but also significantly reduces computation and storage costs, making the evaluation process more efficient and focused.

As soon as a pedestrian enters Area 1, the algorithm recognizes them as Target Pedestrian. This marks the beginning of the risk assessment, during which the algorithm collects data on pedestrian movements and those of nearby vehicles. This process continues until the pedestrian exits Area 3, which indicates that they have safely crossed the intersection and are then considered Nontarget Pedestrian. Although assessing the pedestrian's risk becomes less critical once they enter Area 3 for a proactive protection proposal, the algorithm still completes this step and saves the data for further analysis in post-risk evaluation studies.

When the Target Pedestrian enters Area 2 and gathers enough trajectory data for analysis, totaling 30 data points, the algorithm starts to predict their arrival time at three designated locations, as shown in Figure \ref{fig:Illustration of the Prediction Process}. This prediction process is designed to be adaptive, which means it continuously updates the estimated arrival time based on new trajectory data collected as the pedestrian moves. An essential aspect of this stage involves identifying potential conflict vehicles within each designated conflict area. Specifically, the vehicle that is close to the position of the target pedestrian, especially those in Areas 3 and 4 as outlined in Figure \ref{fig:area-division}, is marked Conflict Vehicle. This approach ensures that situations posing the highest risk are prioritized for evaluation.

Following the estimation of the arrival time for the target pedestrian and surrounding vehicles, the algorithm calculates two Predicted Post-Encroachment Time (P-PET) values for each target pedestrian for each conflict area, as described in Equation \ref{eqn:P-PET_combine}. These P-PET values are crucial to assess the potential risks to pedestrians. The process classifies risks into two categories, Risk 1 and Risk 2, applying a specific P-PET threshold. This helps to differentiate the levels of risk among various groups of pedestrians, such as adults, kids, and cyclists, in two crossing scenarios: when pedestrians cross first or when vehicles cross first. A Target Pedestrian is considered as being at a higher risk level, Risk 2, in a particular conflict area if the count of their P-PET values within the critical threshold exceeds a predetermined limit.

To determine the optimal threshold and counter limit to evaluate the potential risk of the pedestrian, the data set including all pedestrian trajectories extracted in the target area is utilized to explore the best optimal threshold and counter limit. Our approach combines the enumeration technique with a 10-fold cross-validation to find the most effective threshold and counter-limit, aiming to reduce overfitting. We explore a range of threshold values and counter limits, analyzing various combinations to ensure comprehensive coverage. The pair that shows the highest accuracy in the training data set is chosen as the optimal set. To verify that the optimum threshold and counter list is not over-fitted, we further evaluate its performance using four additional metrics on the test dataset as follows.

\begin{itemize}
    \item \textbf{Accuracy}: Reflects the algorithm's overall effectiveness in correctly identifying both Risk 2 and Risk 1 situations.
    \begin{equation}
    \text{Accuracy} = \frac{\text{TP} + \text{TN}}{\text{TP} + \text{TN} + \text{FP} + \text{FN}}
    \end{equation}
    
    \item \textbf{Precision}: Indicates the reliability of the algorithm in identifying actual Risk 2 situations among all detected as Risk 2. Higher precision indicates greater reliability, with fewer Risk 1 situations mistakenly classified as Risk 2, which means lower false alarm rate. 
    \begin{equation}
    \text{Precision} = \frac{\text{TP}}{\text{TP} + \text{FP}}
    \end{equation}
    
    \item \textbf{Recall}: Represents the algorithm's ability to correctly detect all actual high-risk situations. This metric is crucial for the algorithm's effectiveness; a higher recall means fewer high-risk situations go undetected, marking it as a key indicator of algorithm performance.
    \begin{equation}
    \text{Recall} = \frac{\text{TP}}{\text{TP} + \text{FN}}
    \end{equation}
    
    \item \textbf{F1 Score}: Harmonizes precision and recall into a single measure, reflecting both the reliability and effectiveness of the algorithm in evaluating risk.
    \begin{equation}
    \text{F1 Score} = 2 \times \frac{\text{Precision} \times \text{Recall}}{\text{Precision} + \text{Recall}}
    \end{equation}
\end{itemize}

\begin{algorithm}
\caption{Evaluate Pedestrian Risk Using P-PET}
\label{code:Risk}
\resizebox{0.98\textwidth}{!}{ % Start of resizebox
\begin{minipage}{1.3\textwidth} % Adjusted minipage width
\begin{algorithmic}[1]
\Procedure{EvaluatePedestrianRisk}{CCTVVideoDataset}
    \State \textbf{Input:} CCTV video dataset: \textit{Video}
    \State \textbf{Output:} Risk scenario information when target pedestrian is evaluated as Risk2: 
    \begin{equation*}
    \textit{RiskScenarios} = \left\{ \left(ID_{\text{Ped}}, P_t^{ID_{\text{Ped}}}, ID_{\text{Car}}, P_t^{ID_{\text{Car}}}\right)_{1}, \ldots, \left(ID_{\text{Ped}}, P_t^{ID_{\text{Ped}}}, ID_{\text{Car}}, P_t^{ID_{\text{Car}}}\right)_{n} \right\}_{n=1,2,\ldots,N_{\text{total}}}
    \end{equation*}
    
    \textbf{Note:} $\textit{RiskScenarios}$ stores the Pedestrian ID $ID_{\text{Ped}}^{n}$, Pedestrian position $P_t^{ID_{\text{Ped}}^{n}}$ at time $t$, the Conflict Vehicle ID $ID_{\text{Car}}^{n}$, and Conflict Vehicle position $P_t^{ID_{\text{Car}}^{n}}$ at time $t$. $t$ is the moment when the pedestrian is evaluated as risk 2, $N_{\text{total}}$ is the total number of pedestrians evaluated as risk 2 in the \textit{video}.
    \State \textbf{Initialize tools:} Deep SORT, Segment Anything, Transformation Matrix
    \State \textbf{Trained Deep Learning Model Preparation:} Trained YOLOv7, Optimum Arrival Time Prediction Model ${ATPM}_{q}^{i}$
    
    \textbf{Note:} Use \underline{Equation \ref{eqn:Optimum Arrival Time Two Lines}} to select the optimal arrival time prediction model ${ATPM}_{q}^{i}$

    \State \textbf{Predefined Variables:} Area Division, Risk Thresholds [$\alpha_{Q,i}^{s}$, $\beta_{Q,i}^{s}$] for Risk 1 and Risk 2, Counter threshold $\Theta_{Q}$, Minimum Trajectory Length for Prediction $L$
    
    \textbf{Note:} $Q$ represents either the Closer Area or the Further Area, $s$ represents the pedestrian-first or vehicle-first conflict scenario, $i$ represents the category of Target Agent defined in \underline{Section \ref{sec:section3.1.4}}
    \State \textit{RiskScenarios} $\gets$ Initialize empty list
    
    \For{$Frame$ in \textit{Video}}
         \State Utilize Trained YOLOv7, Deep SORT, Segment Anything, Transformation Matrix, Area Division Rules to process $Frame$ to extract trajectory data.
         \State $P_t^i \gets \left\{ (x^i, y^i, ID^i, t)_{1}, \ldots, (x^i, y^i, ID^i, t)_{m} \right\}_{m=1,2,\ldots,M_{\text{total}}}$
         
         \textbf{Note:} $M_{\text{total}}$ represents the total number of observations of the Target Agent $i$ extracted from $Frame$.
        \State Update the trajectory $\text{Traj}_{t-1}^{i,ID}$ by adding the new observation $(x^i, y^i, ID^i, t)$.
        \For{each \textit{pedestrian} with ID $ID_{\text{Ped}}$ in $P_t^i$}
            \If{\textit{pedestrian} \underline{enters} \textbf{Area 1}}
                
        \textbf{Note:} Area is defined by \underline{Section \ref{sec:section3.1.4}}.
                \State Mark \textit{pedestrian} as \textit{TargetPedestrian}
                \State Initialize \textit{RiskCounter} for \textit{TargetPedestrian} at Area $Q$
                \State $RC_{Q}^{ID_{\text{Ped}}}$ $\gets$ 0
            \EndIf
            \If{\textit{TargetPedestrian} \underline{leaves} \textbf{Area 1} and len($\text{Traj}_{t}^{i,ID_{\text{Ped}}}$) $>$ $L$}
                \If{\textit{TargetPedestrian} \underline{leaves} \textbf{Area 3}}
                    \State Change \textit{TargetPedestrian} to \textit{NonTargetPedestrian} 
                    \State Clear \textit{NonTargetPedestrian} Trajectory in $\text{Traj}_{t}^{i,ID_{\text{Ped}}}$ and continue
                \Else
                    \State Use the trained Arrival Time Prediction Model ${ATPM}_{q}^{i}$ to predict arrival time $\hat{\text{ArrT}}_{t}^{i,q}$.
                    \State $\hat{\text{ArrT}}_{t}^{i,q} \gets {ATPM}_{q}^{i}({\text{Traj}_{t-L+1,L}^{i,ID_{\text{Ped}}}})$
                    
                    \textbf{Note:} Only those \textit{TargetPedestrian} in \textbf{Area 2} and \textbf{Area 3} with sufficient trajectory length collected are predicted for arrival time and potential risk is evaluated. ${\text{Traj}_{t-L+1,L}^{i,ID_{\text{Ped}}}}$ represents the sliding window trajectory for agent type $i$ with ID $ID_{\text{Ped}}$ from time $t-L+1$ to time $t$ with length $L$.

                    \For{each conflict area $Q$}
                        \State Mark the vehicle with $ID_{\text{Car}}$ in \textbf{Areas 3} or \textbf{Area 4}, with minimum distance to \textit{TargetPedestrian} as \textit{ConflictVehicle}
                        
                        \If{len$(\text{Traj}_{t}^{i,ID_{\text{ConflictVehicle}}})$ $\geq$ L}
                            \State Use the trained Arrival Time Prediction Model ${ATPM}_{q}^{i}$ to predict the arrival time $\hat{\text{ArrT}}_{t}^{i,q}$ for ConflictVehicles
                            \State Calculate ${P-PET}_{VF,Q}$, ${P-PET}_{PF,Q}$ values using \underline{Equation \ref{eqn:P-PET_combine}}
                            \If{P-PET $\in$ [$\alpha_{Q,i}^{s}$, $\beta_{Q,i}^{s}$]}
                                \State $RC_{Q}^{ID_{\text{Ped}}}$ $\gets$ $RC_{Q}^{ID_{\text{Ped}}}$ + 1
                            \EndIf
                        \EndIf
                        \If {$RC_{Q}^{ID_{\text{Ped}}}$ $>$ $\Theta_{Q}$}
                            \State Classify \textit{TargetPedestrian} as Risk 2
                            \State \textit{RiskScenarios} add $\left(ID_{\text{Ped}}, P_t^{ID_{\text{Ped}}}, ID_{\text{Car}}, P_t^{ID_{\text{Car}}}\right)$
                            \State \underline{Protection Algorithm} Perform
                            
                            \textbf{Note:} Once the \textit{TargetPedestrian} is classified as Risk 2, the designed \underline{Protection Algorithm} (not in this study) can be immediately performed to protect the pedestrian.     
                        \EndIf
                    \EndFor
                \EndIf
            \EndIf    
        \EndFor
    \EndFor
    \State \textbf{return} \textit{RiskScenarios}
\EndProcedure
\end{algorithmic}
\end{minipage} % End of minipage
} % End of resizebox
\end{algorithm}
\newpage

\section{Result}
\label{sec:section 3}
\subsection{Dataset Extraction}
\label{sec:section 3.1}
%main article
\subsubsection{Detection Performance}

The detection results of our trained YOLOv7 model, as shown in Table \ref{table:detection model}, demonstrate the model's ability to differentiate among various categories of pedestrians: kids, adults, and cyclists. A total of 1337 pedestrians were labeled as instructed in Section \ref{sec:section3.1.3}, serving as the ground truth to evaluate the performance of the trained detection model. In particular, the model shows exceptional precision in adult classification, showing its ability to accurately identify adult pedestrians. In contrast, the precision for detecting kids and cyclists is lower, pointing to a slightly higher rate of misclassification for these pedestrian groups. In general, this result highlights the overall effectiveness of the model in pedestrian detection, while also indicating that it is more challenging for the classifications of kids and cyclists.

\begin{table}[ht]
\centering
\resizebox{0.8\textwidth}{!}{%
\begin{tabular}{lcccccc}
\hline
\textbf{Types} & \textbf{True Positive} & \textbf{False Positive} & \textbf{False Negative} & \textbf{Precision} & \textbf{Recall} & \textbf{F1 Score} \\
\hline
Kids & 125 & 30 & 19 & 0.806 & 0.868 & 0.836 \\
Adult & 820 & 44 & 61 & 0.949 & 0.931 & 0.939 \\
Cyclists & 278 & 39 & 33 & 0.877 & 0.894 & 0.885 \\
\hline
\end{tabular}}
\caption{Performance Metrics of Custom-Trained YOLOv7 Detection Model on Pedestrian Classification: Kids, Adult, and Bike Categories.}
\label{table:detection model}
\end{table}

\subsubsection{Crossing Time Distribution}

\begin{figure}[!ht]
  \centering
  \includegraphics[width=0.6\textwidth]{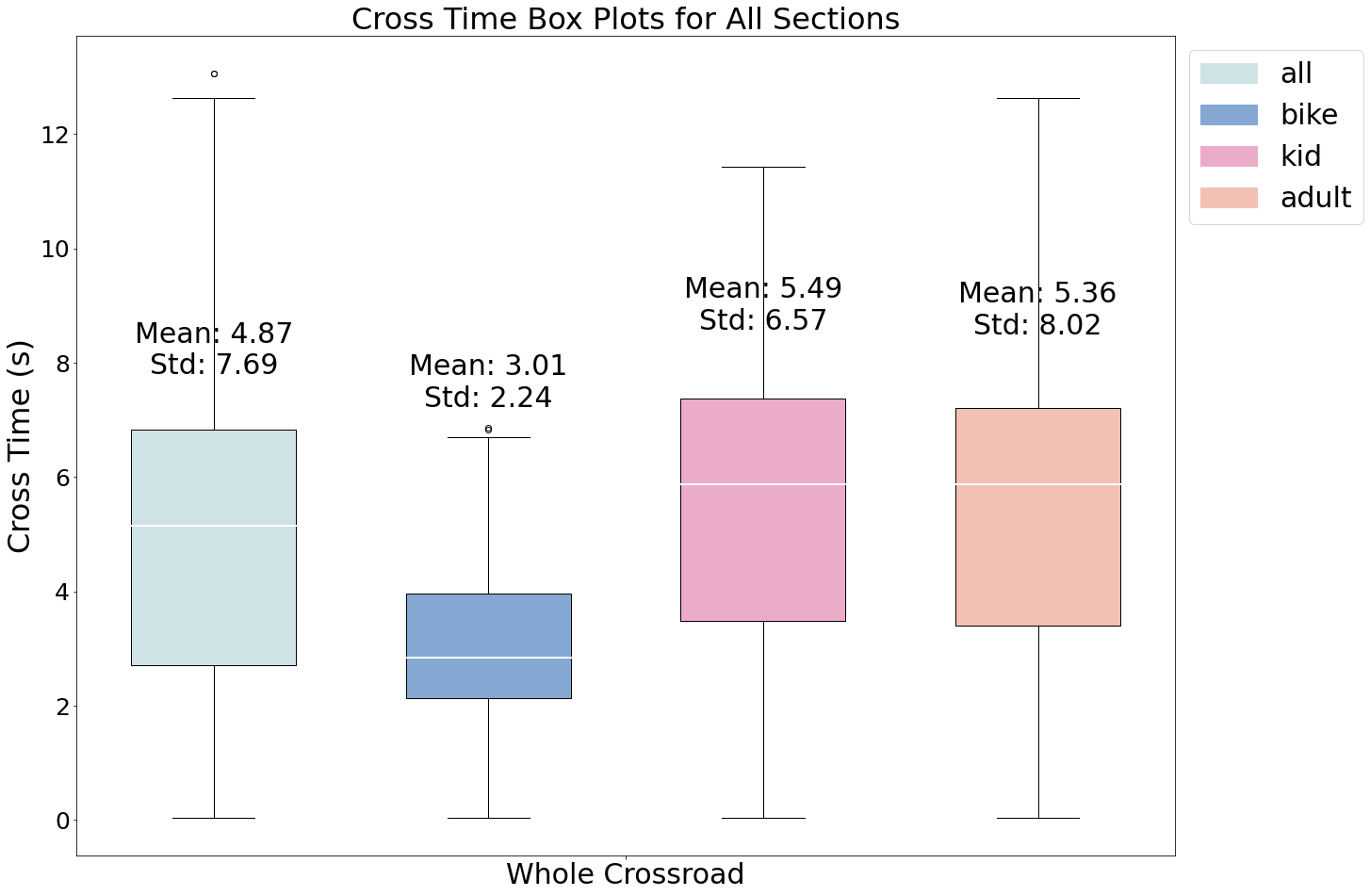}
  \caption{Boxplot of Intersection Crossing Times by Pedestrian Type}\label{fig:Behavior of different types of pedestrian}
\end{figure}

Figure \ref{fig:Behavior of different types of pedestrian} illustrates a box plot that compares the crossing time of kids, adults, and cyclists. It reveals that adults and kids exhibit similar crossing time with means of 5.49 seconds and 5.36 seconds, respectively. Cyclists are the fastest across the intersection, with a mean time of 3.01 seconds and the least variation in time.

\begin{figure}[!ht]
  \centering
  \includegraphics[width=0.6\textwidth]{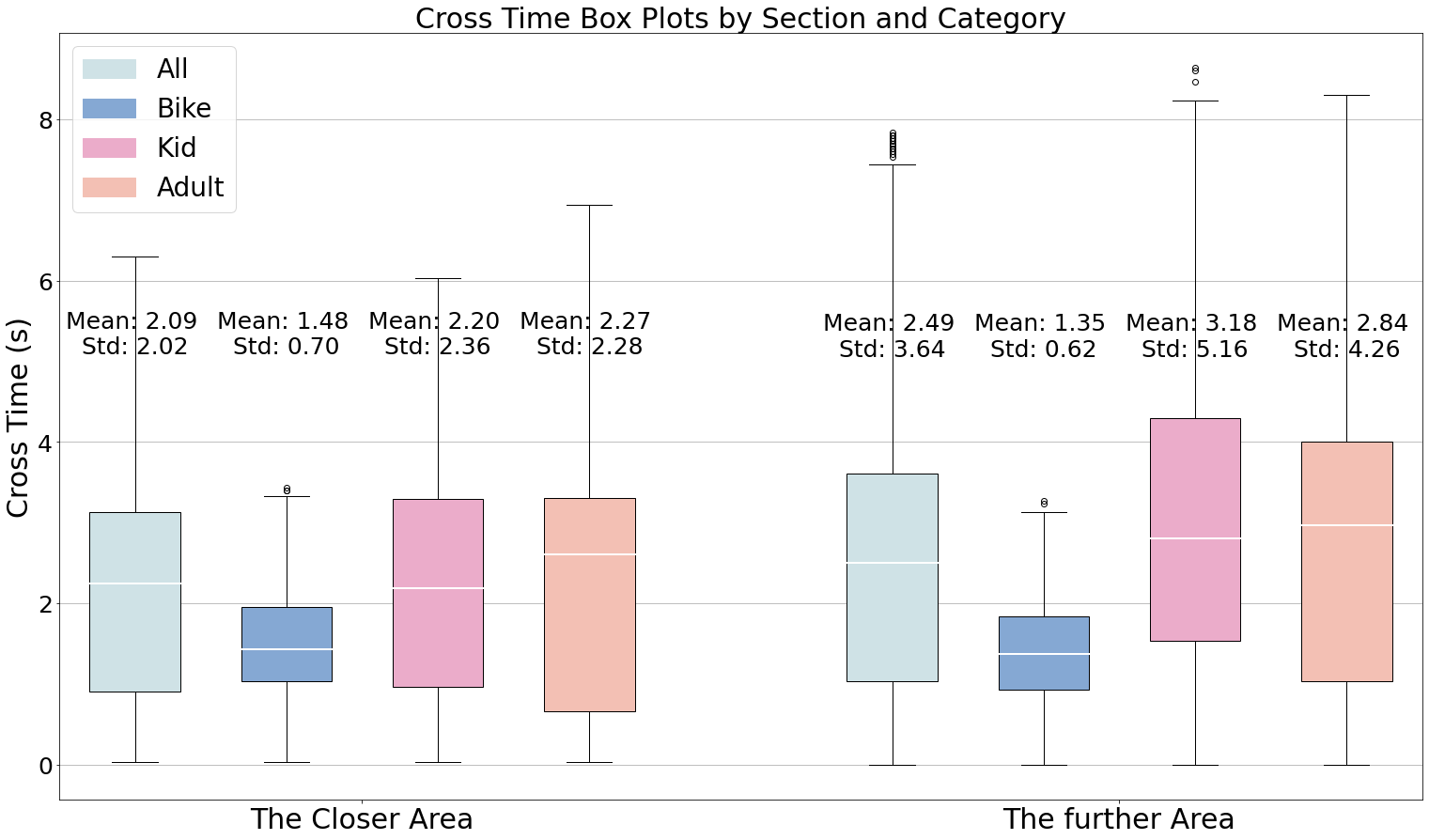}
  \caption{Boxplot of Intersection Crossing Times for Pedestrians in Closer and Further Areas}\label{fig:Behavior of different types of pedestrian at two area}
\end{figure}

The box plot in Figure \ref{fig:Behavior of different types of pedestrian at two area} offers a detailed analysis of the crossing behaviors of various types of pedestrians on different sections of the road. Cyclists are consistently the fastest, kids demonstrate a relatively rapid mean crossing time of 2.20 seconds in the closer area, yet exhibit a longer mean time of 3.18 seconds in the further area, with a notable standard deviation of 5.16 seconds. This high variability in kids' crossing time, particularly in the further area, confirms the common understanding that kids' movements can be quite unpredictable. Adults, with a mean crossing time of 2.27 seconds in the closer area and 2.84 seconds in the further area, show less variability than kids but still a wide range of crossing time, especially in the further area. Cyclists maintain the lowest variability, indicative of their consistent and rapid crossing through the intersection. The results emphasize the importance of categorizing different types of pedestrian in specific areas for the performance of the pedestrian's potential risk evaluation.

\subsection{Future State Prediction}
\label{sec:section 3.2}
\subsubsection{Dataset Constitution}
In the future state prediction, we focus on the increased risk faced by pedestrians who do not notice approaching vehicles, operating under the assumption that these individuals are at higher risk than those who have noticed the presence of vehicles. Therefore, our study uses the dataset consisting only of pedestrians who did not notice oncoming vehicles for the model training. This specific focus is used to train our arrival time prediction model, ensuring that it can predict the future state in a higher-risk situation.

\begin{figure}[!ht]
  \centering
  \includegraphics[width=\textwidth]{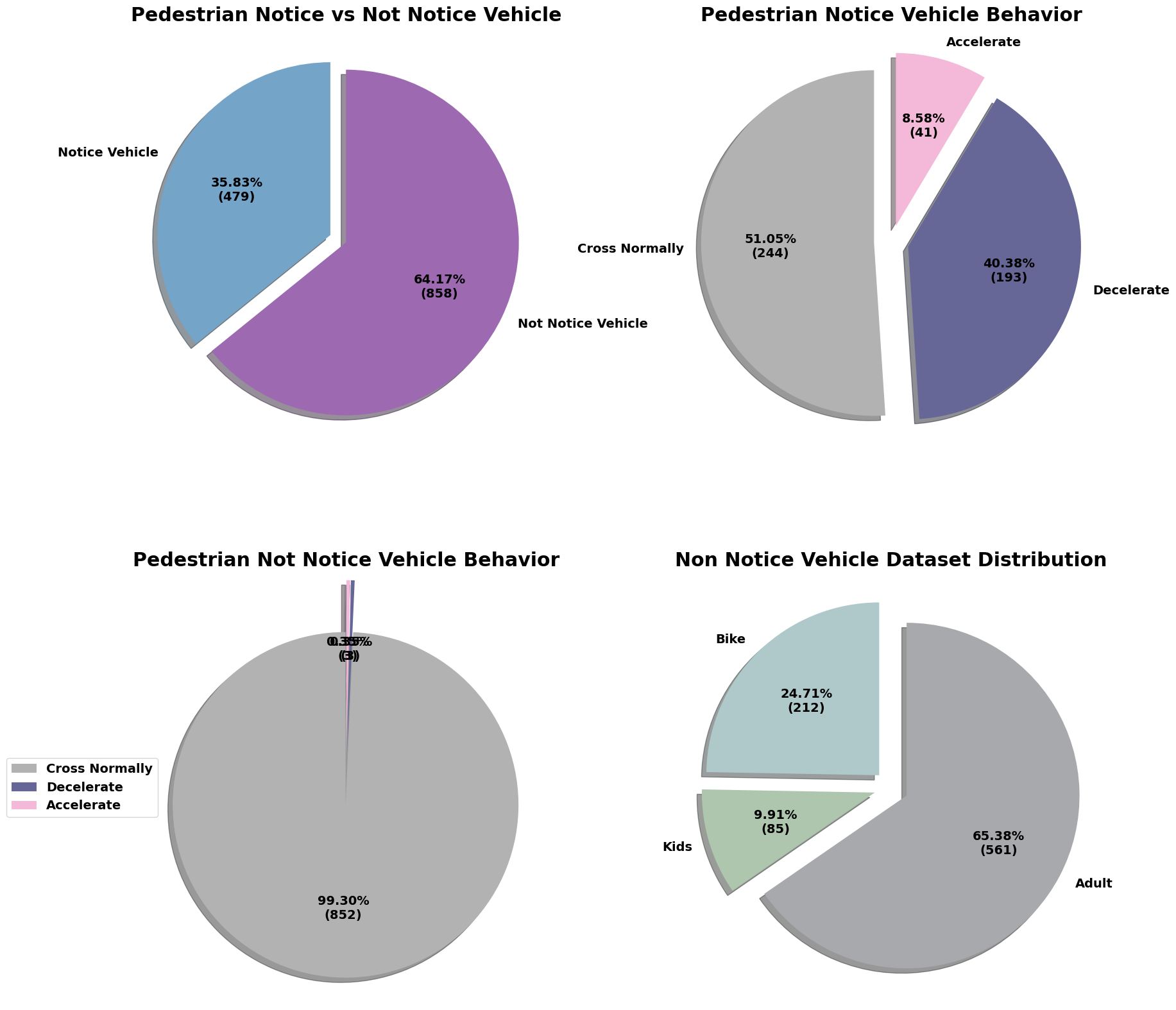}
  \caption{Composition of Dataset and Behaviors of Pedestrians Noticing vs. Not Noticing Vehicles}\label{fig: Data Constitude & Pedestrain Behavior}
\end{figure}

The constitution of our dataset, as presented in Figure \ref{fig: Data Constitude & Pedestrain Behavior}, reveals the various behaviors exhibited by different types of pedestrians when confronted with approaching vehicles. Although 64.17\% of pedestrians do not notice vehicles, potentially increasing their risk of accidents due to inattention, the remaining 35.83\% who notice vehicles mostly continue to cross normally, implying a perceived sense of safety or right of way. Among pedestrians aware of the vehicles, a significant portion decelerates, indicating a cautious approach, while only a few accelerate to quickly leave the danger area. In terms of concern, the vast majority of those who do not notice vehicles (99.30\%) cross normally without prevented reaction to their surroundings, highlighting the considerable risk of unnoticed vehicle-pedestrian interactions.

The dataset distribution for non-noticing individuals skews heavily toward adults, who constitute 65.38\%, followed by cyclists at 24.71\%. Kids, while the smallest group, 9.91\% who did not notice the vehicle, still represents a particularly vulnerable pedestrian due to the potential lack of awareness of road safety and visibility issues.

\subsubsection{Arrival Time Prediction}

\begin{table}[]
\centering
\caption{Performance Metrics Comparison: 
All Pedestrians (Noticed and Non-noticed) vs. Those didn't notice the vehicles}
\label{table:all_and_no_notice}
\scriptsize
\resizebox{0.9\textwidth}{!}{%
\begin{tabular}{l|ccc|ccc}
\hline
            & \multicolumn{3}{c|}{All Data}                                             & \multicolumn{3}{c}{No Notice Data}                                      \\
location            & \multicolumn{1}{c}{q=0}    & \multicolumn{1}{c}{q=1}    & \multicolumn{1}{c|}{q=2}    & \multicolumn{1}{c}{q=0}    & \multicolumn{1}{c}{q=1}    & \multicolumn{1}{c}{q=2}    \\
Model       & RMSE / MAE & RMSE / MAE & RMSE / MAE & RMSE / MAE & RMSE / MAE & RMSE \\ \hline
HA          & 9.708 / 3.059 & 10.59 / 5.36 & 15.54 / 8.490 & 2.677 / 1.486 & 4.971 / 3.144 & 11.7 / 6.504 \\
LSTM        & 2.009 / 1.445 & 2.471 / 1.895 & 4.161 / 3.449 & 1.59 / 1.128 & 2.042 / 1.497 & 2.482 / 1.795 \\
GRU         & 2.04 / 1.458 & 2.541 / 1.963 & 4.64 / 3.95 & 1.724 / 1.2 & 2.251 / 1.672 & 2.584 / 1.874 \\
Transformer & 2.278 / 1.753 & 2.649 / 2.042 & 3.648 / 2.924 & 1.804 / 1.449 & 2.186 / 1.654 & 2.923 / 2.318 \\ \hline
\end{tabular}}
\end{table}
In this section, we evaluate the performance of the arrival time prediction models using two standard metrics for assessing prediction accuracy: the Root Mean Square Error (RMSE) and the Mean Absolute Error (MAE). While RMSE gives us a measure of the average magnitude of the error, providing a sense of how far the predictions typically deviate from the actual values, MAE offers a linear score that reflects the average absolute difference, thus presenting a more direct interpretation of the average error magnitude.

\paragraph{Impacts on Vehicle Notification}
Table \ref{table:all_and_no_notice} compares the performance of different models using a complete data set (including pedestrians who notice and do not notice vehicles) and a subset of data focused on those who do not notice vehicles. 
The results of model performance across 'All Data' and 'No Notice Data' showcase distinct variations in prediction accuracy. 
The RMSE and MAE values for the 'No Notice Data' dataset consistently surpass those for 'All Data', emphasizing the challenges deep learning models encounter in forecasting pedestrian behaviors when it involves subjective reactions, such as deceleration or acceleration, to avoid approaching vehicles. Therefore, based on this understanding and the results observed, we predict the arrival time under the assumption that all pedestrians are in a higher-risk scenario akin to the 'Not Notice Vehicle' condition. This assumption allows us to focus on the worst-case scenario in risk evaluation, ensuring that predictive models are tailored toward increased caution and prioritizing safety in potential high-risk interactions between pedestrians and vehicles.

\paragraph{Pedestrian Class Impacts}
As outlined in Section 4.1.2, there are notable differences in the behaviors of adults, kids, and cyclists. These distinct behaviors are reflected in the predictive model performance metrics detailed in Table \ref{table_combined}. For evaluation, prediction models were specifically trained using datasets that exclusively contain instances of each type of pedestrian.  RMSE and MAE metrics reveal that kids consistently have the highest prediction errors in all target areas, implying that their movement patterns may be more unpredictable or variable compared to cyclists and adults. On the contrary, cyclists show the lowest error rates, likely due to their relatively uniform speeds and direct trajectories. Adults show intermediate error levels, suggesting that their movements strike a middle ground of predictability and variability in speed and path choices.

\begin{table}[]
\centering
\caption{Performance Metrics for Kids, Cyclists, and Adults}
\label{table_combined}
\resizebox{\textwidth}{!}{%
\begin{tabular}{l|ccc|ccc|ccc}
\hline
Catagories    & \multicolumn{3}{c|}{Kids} & \multicolumn{3}{c|}{Cyclists} & \multicolumn{3}{c}{Adults} \\
location    & q=0        & q=1        & q=2       & q=0         & q=1         & q=2        & q=0         & q=1         & q=2        \\ 
Model       & RMSE / MAE  & RMSE / MAE  & RMSE / MAE & RMSE / MAE    & RMSE / MAE    & RMSE / MAE  & RMSE / MAE    & RMSE / MAE    & RMSE / MAE  \\ \hline
HA          & 2.006/1.533 & 4.516/4.040 & 21.54/10.11& 1.611/1.099   & 2.098/1.892    & 5.013/4.360 & 2.492/1.513    & 3.997/3.322  & 10.45/6.767 \\
LSTM        & \textbf{1.791/1.456} & \textbf{2.293/1.813} & \textbf{2.279/1.762} & 0.8989/0.7088 & \textbf{1.166/0.8887}  & \textbf{1.578/1.286} & \textbf{1.175/0.8729}  & \textbf{1.818/1.457}   & \textbf{2.128/1.639} \\
GRU         & 2.359/1.662 & 2.956/2.264 & 3.41/2.7    & \textbf{0.8345/0.609} & 1.252/0.9786  & 1.952/1.602 & 1.207/0.922   & 1.94/1.597    & 2.389/1.914 \\
Transformer & 2.263/1.962 & 3.613/3.093 & 4.661/3.927 & 1.054/0.8255  & 1.484/1.211   & 2.655/2.239 & 1.217/0.9269  & 1.97/1.669    & 2.967/2.459 \\ \hline
\end{tabular}}
\end{table}

\paragraph{Results on Historical Average Model}
Analyzing the performance metrics from Tables \ref{table:all_and_no_notice} and \ref{table_combined}, it becomes clear that the historical average (HA) model, although it is a traditionally preferred method to estimate arrival time, demonstrates significantly higher errors compared to deep learning models such as LSTM, GRU, and Transformer. Particularly in the context of q = 2, which represents the further area from the point of origin in the trajectory, the error margins of the HA model (RMSE / MAE) contrast starkly with the more advanced methods. For example, in the 'All Data' scenario for location q = 2, the HA model shows an RMSE of 15.54 and an MAE of 8.490, considerably higher than those observed for deep learning models. This trend persists whether we evaluate all pedestrians or focus solely on those who do not notice vehicles. Given the substantial errors associated with HA, particularly in critical scenarios for the assessment of pedestrian safety, the adoption of deep learning methods offers a compelling alternative to improve the accuracy in predicting pedestrian risk.

\paragraph{Different Conflict Area Impacts}
The comparison of model performances across different target locations (q = 0, q = 1, and q = 2) as presented in the tables highlights a general trend: Predicting pedestrian arrival time becomes increasingly challenging as the target location moves further from the target pedestrian's current position. Specifically, q = 0, which is closer to the pedestrian starting point, consistently shows lower RMSE and MAE values in all models and data subsets, suggesting that predictions in the immediate vicinity of the pedestrian's origin are more accurate. This can be attributed to the reduction of complexity and shorter prediction horizons, which produce less uncertainty. As we progress to q = 1 and especially to q = 2, the error rates increase markedly, underscoring the complexity of longer-term predictions, where factors such as pedestrian speed variations, vehicle interactions, and potential direction changes contribute to the difficulty of prediction. 

\paragraph{Best Performance Model for Pedestrian}
Analyzing the performance metrics in Table \ref{table_combined} provides clear information on the predictive models that excel under various conditions to predict the arrival time for kids, cyclists and adults in different zones. The LSTM model demonstrates notable proficiency, especially for kids and adults, by achieving the lowest RMSE and MAE values in most cases. For cyclists, characterized by more consistent and predictable behavior, the GRU model excels in close range (q = 0), demonstrating its ability to capture regular motion patterns. However, the LSTM model maintains a slight advantage at intermediate distances (q = 1) and longer (q = 2) for all types of pedestrians. While the Transformer model is often hailed for its overall superior performance in multiple domains, it falls short in this particular scenario due to the limited size of the dataset. Our analysis suggests that the LSTM model is more suitable for predicting arrival time with limited data.

\begin{table}[]
\centering
\caption{Performance Metrics for Cars in different Areas}
\label{table:prediction_for_car}
\resizebox{0.7\textwidth}{!}{%
\begin{tabular}{l|cc|cc|cc}
\hline
            & \multicolumn{2}{c|}{Area 4.1} & \multicolumn{2}{c|}{Area 4.2} & \multicolumn{2}{c}{All Area} \\
Location      & q = 0           & q = 1           & q = 0            & q = 1            & q = 0          & q = 1         \\ \hline
HA          & 3.294/2.264    & 7.794/5.687   & 16.68/6.987    & 25.91/9.802    & -/-         & -/-        \\
LSTM        & 1.653/1.148    & 2.234/1.694   & \textbf{1.221/0.9386}   & \textbf{3.205/2.936}    & 2.216/1.575  & 2.712/2.114 \\
GRU         & \textbf{1.591/1.111}    & \textbf{2.169/1.641 }  & 1.185/0.9591  & 3.48/3.22      & 2.177/1.614  & 3.206/2.596 \\
Transformer & 1.922/1.427    & 2.255/1.68    & 1.517/1.21     & 3.366/3.371    & 2.714/2.038  & 3.163/2.569 \\ \hline
\end{tabular}}
\end{table}

\paragraph{Model Performance for Vehicle}

Table \ref{table:prediction_for_car} highlights the performance of various models in predicting vehicle behaviors in two distinct zones: Areas 4.1 and 4.2, together with the aggregate performance for all areas of the car. This comparison emphasizes the advantage of tailoring predictive models to distinct areas to improve accuracy. When observing the performance of the models, it is evident that for the location entering the conflict area (q = 0), both areas exhibit commendable accuracy across the models. However, since the prediction extends to further distances (q = 1), particularly in Area 4.2, there is a notable increase in error. This decreased performance in Area 4.2 for q = 1 could be attributed to its more complex layout, characterized by curved trajectories of vehicles approaching from multiple directions.

\subsection{Potential Risk Evaluation}
\label{sec:section 3.3}
\subsubsection{Dataset Component}

\begin{figure}[!ht]
  \centering
  \includegraphics[width=0.9\textwidth]{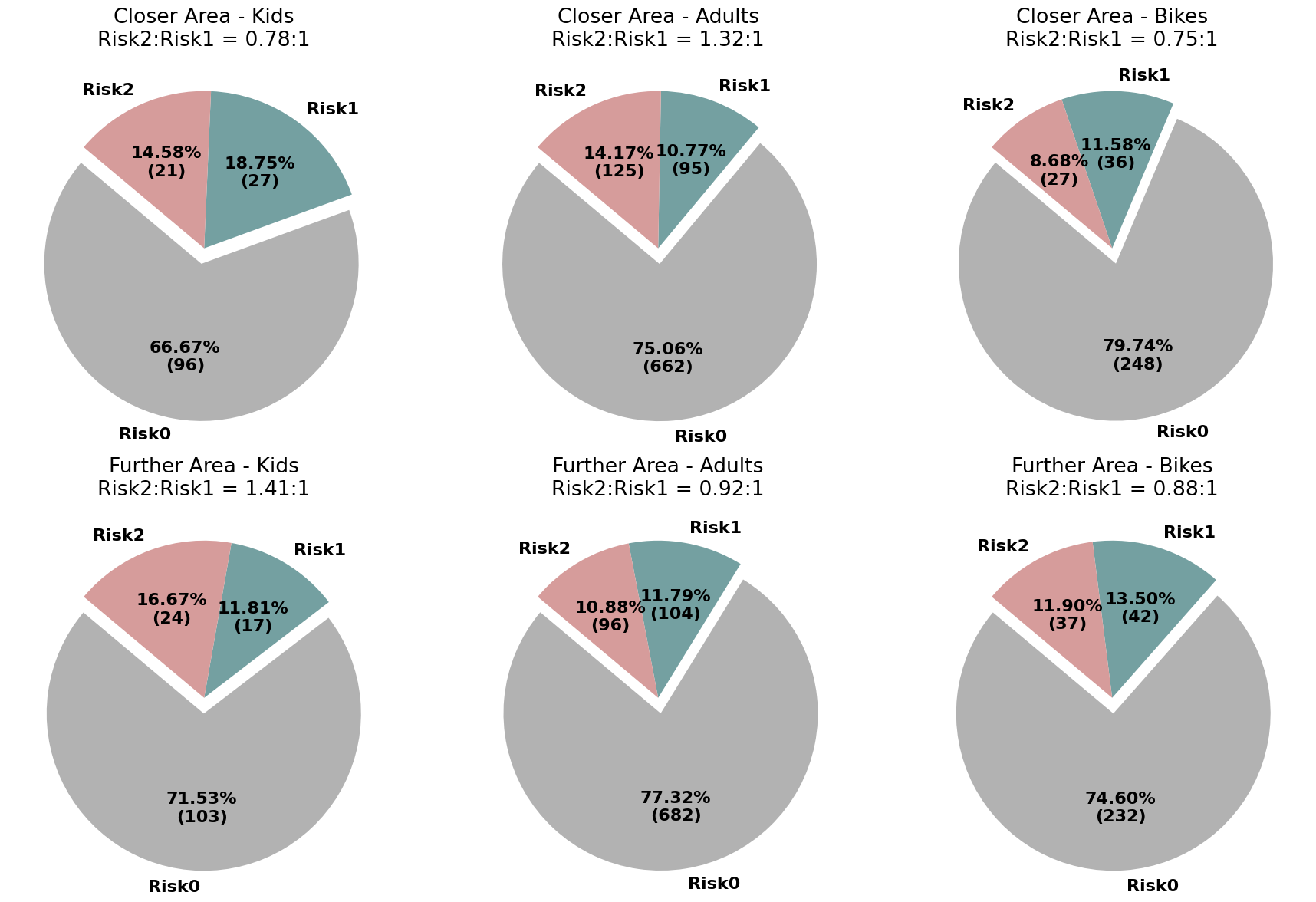}
  \caption{Risk Composition Analysis for Different Pedestrian Types: Kids, Adults, and Cyclists}\label{fig: Risk Evaluation consittude}
\end{figure}

The pie chart presented in Figure \ref{fig: Risk Evaluation consittude} emphasis distinct safety dynamics among kids, adults, and cyclists within varying proximity to vehicular traffic. The analysis reveals that kids exhibit a lower Risk2: Risk1 ratio in the Closer Area (= 0.78), but a significantly higher ratio in the Further Area (= 1.41), suggesting a worrying increase in high-risk situations as they distance themselves from immediate traffic threats. On the contrary, adults demonstrate a higher propensity for high-risk encounters in the Closer Area (= 1.32) with a notable reduction in the Further Area (= 0.92), indicating an improved ability to manage safety with increased distance from traffic. Cyclists maintain the lowest Risk2: Risk1 ratios in both areas (= 0.75 for Closer Area and = 0.88 for Further Area), reflecting their consistent vigilance and ability to mitigate high-risk scenarios. This illuminates the critical need for tailored safety interventions.

It is important to note that the size of the data set for kids and cyclists is relatively small, which could limit the ability to definitively identify patterns. To address potential biases arising from the dataset's size, we employ a ten-fold cross-validation method. This approach improves the robustness of the threshold determined for risk evaluations, ensuring that it remains reliable and is not overly adapted to the constrained dataset.

\subsubsection{Algorithm based on P-PET Evaluation}
\begin{table}[]
\centering
\caption{Risk Evaluation Algorithm Performance Comparison in Closer and Further Areas}
\label{table:Combined Risk Evaluation Algorithm Performance}
\resizebox{\textwidth}{!}{%
\begin{tabular}{c|ccccccc|ccccccc}
\hline
Area & \multicolumn{7}{c|}{Closer Area} & \multicolumn{7}{c}{Further Area} \\
Category  & Test Acc. & Precision & Recall & F1 Score & PF\_PPET                 & CF\_PPET               & Counter & Test Acc. & Precision & Recall & F1 Score & PF\_PPET                 & CF\_PPET               & Counter \\ \hline
W/o class & 0.7706    & 0.7333    & 0.8846 & 0.8000   & $[-1.0, -0.5]$ & $[0.0, 1.0]$ & 3       & 0.7531    & 0.7829    & 0.6859 & 0.7272 & $[-3.0, -1.5]$ & $[2.0, 2.5]$ & 3       \\ \hline
W/ class  &  0.8188   &  0.7884   &  0.8549 &  0.8023   & -                        & -                      & -       &  0.7991    &  0.8279    &  0.7260 &  0.7534   & -                        & -                      & -       \\
\textbf{Adult}     & \textbf{0.8000}    & \textbf{0.7774}    & \textbf{0.9178} & \textbf{0.8350}   & \textbf{[-0.7, 0.1]}  & \textbf{[0.1,1.1]}  & \textbf{3}       & \textbf{0.7900}    & \textbf{0.8092}    & \textbf{0.7389} & \textbf{0.7607}   & \textbf{[-2.5, -1.5]} & \textbf{[0.9, 2.4]} & \textbf{3}       \\
Kid       & 0.8700    & 0.9000    & 0.6500 & 0.7300   & $[-1.8,-1.6]$  & $[0,0.8]$    & 3       & 0.7800    & 0.7000    & 0.6166 & 0.6367   & [-3.0,-0.2]  & [-1.6,0.8] & 3       \\
\textbf{Cyclist}      & \textbf{0.8452}    & \textbf{0.7417}    & \textbf{0.7917} & \textbf{0.7431}   & \textbf{[-3.0,-2.6]}  & \textbf{[1.4,2.0]}  & \textbf{5}       & \textbf{0.8321}    & \textbf{0.9417}    & \textbf{0.7500} & \textbf{0.7957}   & \textbf{[-0.6,0.2]}   & \textbf{[0.6,1.6]}  & \textbf{3 }      \\ \hline
\end{tabular}}
\end{table}

%% with class without class

%% adult kid bike

%% closer further
%% with area without area

\begin{table}[]
\centering
\caption{Risk Evaluation Algorithm Performance w/ and w/o Area Division}
\label{table:Combined Risk Evaluation Algorithm Performance w/ and w/o Area Division}
\resizebox{0.8\textwidth}{!}{%
\begin{tabular}{c|ccccccc}
\hline
Category                   & Test Acc. & Precision & Recall & F1 Score & PF\_PPET         & CF\_PPET       & Counter \\ \hline
W/o Area Division             & 0.7347    & 0.7260    & 0.8334 & 0.7657   & -                & -              & -       \\
Adult                      & 0.7218    & 0.7191    & 0.8364 & 0.7692   & {[}-1.0, -0.5{]} & {[}0.0, 0.5{]} & 3       \\
\textbf{Kid}                & \textbf{0.7464}    & \textbf{0.7233}    & \textbf{0.8767} & \textbf{0.7543}   & \textbf{[-3.3,-3.1]}  & \textbf{[1.0, 1.5]} & \textbf{3}       \\
Cyclist                       & 0.7750    & 0.7837    & 0.7908 & 0.7720   & {[}-1.6, -1.2{]} & {[}0.7, 1.2{]} & 4       \\ \hline

\end{tabular}}
\end{table}

The performance evaluation of the potential risk assessment Algorithm \ref{code:Risk}, outlined in Tables \ref{table:Combined Risk Evaluation Algorithm Performance} and \ref{table:Combined Risk Evaluation Algorithm Performance w/ and w/o Area Division}, employs optimally pretrained arrival time prediction models. This comprehensive analysis demonstrates the algorithm's ability to evaluate pedestrian's potential risk within the closer and further conflict area. In this study, four different evaluation matrices are compared, namely accuracy, precision, recall, and F1 score, as introduced in Section \ref{sec:section3.3.2}.

\paragraph{Different Conflict Area Impact}
The algorithm's performance demonstrates clear variations between the closer and further conflict areas, as detailed in Table \ref{table:Combined Risk Evaluation Algorithm Performance}. There is a noticeable decrease in performance at further area, a phenomenon consistent with different categories of pedestrians, such as adults and kids. This variation highlights the intrinsic challenge of accurately predicting risk over long distances. Of particular note is the performance of cyclists in the further area: Despite the high precision, the recall rate drops from 79.17\% to 75\%. This pattern implies that while the algorithm is adept at classifying many scenarios as non-risky(Risk 1), it has a propensity to miss some critical high-risk (Risk 2) scenarios at these greater distances, which leads to a lower effectiveness of the algorithm in the further distance.

\paragraph{Classification Impact}
The necessity for distinct thresholds for various classes of pedestrians becomes particularly clear when comparing the performance results "with classification" with those "without classification" in Table \ref{table:Combined Risk Evaluation Algorithm Performance}. Data unmistakably show that classifying pedestrians into groups such as adults, kids, and cyclists and applying tailored thresholds yields similar performance in the closer distance, but significant improvement in the further distance. 

Although the model achieves high test precision to assess risk in kids, its recall rates are relatively lower. This suggests that although the overall accuracy is high, the model's sensitivity to detect high-risk situations involving kids should be improved. The unpredictable nature of the movements of the kids necessitates an enhancement in the algorithm's sensitivity to ensure that all potential high-risk scenarios are accurately identified, highlighting the need for further refinement to better protect this vulnerable group.

To improve the algorithm's sensitivity to high-risk scenarios for kids, we implemented a single threshold range applicable for both closer and farther areas, moving away from the method of setting unique thresholds for each area. This consolidated strategy, by aggregating Predicted Post-encroachment Time (P-PET) measurements from both nearer and further distances, designates a scenario as high-risk (Risk 2) when the aggregate P-PET counts within the specified range across both regions surpass a unified threshold. Designed to enhance the algorithm's sensitivity to recognize high-risk situations, this strategy may slightly increase the false detection rate. The results of this refined method are elaborated in Table \ref{table:Combined Risk Evaluation Algorithm Performance w/ and w/o Area Division}. Data indicate an improvement in recall for kids: from 65\% in the closer areas and 61.66\% in the farther areas to 83.17\% in the new settings despite a minor decrease in test precision and precision. For adults and cyclists, the benefits of improving the F1 score are less pronounced, suggesting that an area-specific threshold could be more suitable for these groups. Consequently, as shown in Table \ref{table:Overall Performance of Risk Evaluation Algorithm}, the refined risk evaluation algorithm achieves a recall rate of 82.64\% and an F1 score of 78.09\%, indicating its effectiveness in identifying potential risks.

\paragraph{Threshold of Two P-PET Scenario}
The values of ${P-PET}_{PF}$ and ${P-PET}_{CF}$ presented in Tables \ref{table:Combined Risk Evaluation Algorithm Performance} and \ref{table:Combined Risk Evaluation Algorithm Performance w/ and w/o Area Division}, derived from our analysis, support our approach of using Predicted Post-Encroachment Time (P-PET) for evaluating pedestrian risk at intersections. Specifically, the values of ${P-PET}_{PF}$ fall within a small negative range, indicating a significant probability that a pedestrian will be hit by a vehicle from the front in scenarios where the pedestrian crosses first. Regarding ${P-PET}_{CF}$, the values range from small negative to small positive, which aligns with our expectations. Negative values hint at a potential frontal collision, while positive values signal scenarios where the vehicle occupies the crosswalk, both indicating high-risk situations. The ranges ${P-PET}_{PF}$ and ${P-PET}_{CF}$ align with our initial assumptions, strengthening the validity of our method of risk evaluation.

\begin{table}[]
\centering
\caption{Overall Performance of Risk Evaluation Algorithm }
\label{table:Overall Performance of Risk Evaluation Algorithm}
\resizebox{0.450\textwidth}{!}{%
\begin{tabular}{c|cccc}
\hline
Category            & Test Acc. & Precision & Recall & F1 Score \\ \hline
\textbf{Overall}            & \textbf{0.7897}    & \textbf{0.7801}    & \textbf{0.8264} & \textbf{0.7809}   \\ \hline
Adult               & 0.7952    & 0.7926    & 0.8326 & 0.7996   \\
Kid                 & 0.7464    & 0.7233    & 0.8767 & 0.7543   \\
Bike                & 0.8390    & 0.8369    & 0.7718 & 0.7682   \\ \hline
\end{tabular}}
\end{table}

\subsubsection{Real-Time Feasibility}
The real-time requirement is crucial to evaluate the potential risk to pedestrians, ensuring that proactive safety measures can be implemented practically. Given the nature of video flow at 30 frames per second (FPS), it is imperative that each component of the algorithm completes its execution in 33 milliseconds to maintain real-time processing. This constraint highlights the importance of optimizing each function within the algorithm to meet the demands of real-time performance and response in dynamic pedestrian environments.

For our experimental setup, we utilized high-performance computing devices to assess the theoretical feasibility of real-time algorithm implementation. GPU-based experiments were conducted on an NVIDIA RTX 3090 Ti, equipped with 24 GB of VRAM, known for its superior processing capabilities and efficiency in handling complex computational tasks. This GPU is particularly suited for demanding tasks such as object detection, tracking, and segmentation, which are integral to our algorithm. On the CPU front, we employed an AMD Ryzen 9 5900X 12-Core Processor, renowned for its high throughput and multitasking performance. Considering the impracticality of installing high-computation GPUs in CCTV systems due to the physical size limitation, our exploration extends to the practical feasibility of utilizing edge devices for processing with the mentioned CPU.

\begin{table}[ht]
\centering
\caption{Detailed Average Time Costs for Each Algorithm Component Across Devices}
\label{table:algorithm_performance}
\resizebox{0.85\textwidth}{!}{% Adjust the resizebox if necessary to fit your page layout
\begin{tabular}{l l c c c l}
\hline
\textbf{Algorithm Category}          & \textbf{Algorithm Function} & \textbf{Average Time [ms]} & \textbf{Std [ms]} & \textbf{Unit} & \textbf{Device} \\ \hline
\multirow{5}{*}{Computer Vision}     & Object Detection            & 25.87                      & 2.536             & Frame         & GPU             \\
                                     &                             & 670.8                      & 21.75             & Frame         & CPU             \\
                                     & Object Tracking             & 12.85                      & 5.474             & Frame         & GPU             \\
                                     &                             & 11.43                      & 30.82             & Frame         & CPU             \\
                                     & Segmentation                & 489.8                      & 55.42             & ID            & GPU             \\ \hline
\multirow{3}{*}{Safety Evaluation}   & \textbf{Arrival Time Prediction}     & \textbf{5.341}             & \textbf{2.95}     & \textbf{ID }           & \textbf{GPU}    \\
                                     &                             & 31.02                      & 11.39             & ID            & CPU             \\
                                     & \textbf{Potential Risk Evaluation}   & \textbf{1.516}             & \textbf{0.5067}   & \textbf{ID}            & \textbf{CPU}    \\ \hline
\end{tabular}}
\end{table}

Detailed analysis of the performance of the framework functions on different devices, as captured in our experimental setup, shown in Table \ref{table:algorithm_performance}, reveals significant insights into the real-time feasibility of implementing our comprehensive pedestrian risk evaluation system. Specifically, the Arrival-Time Prediction and Potential Risk Evaluation algorithms show promising results, with execution time well within the real-time requirement of 33 milliseconds per frame, particularly when run on a GPU. Notably, the Arrival Time Prediction algorithm demonstrates remarkable efficiency with an average processing time of approximately 5 ms on the GPU, although this increases significantly to around 31 ms on the CPU.

The segmentation function, while beneficial for refining pedestrian trajectory data and reducing the error from bounded box, poses a considerable challenge due to its high computational demand, evident from its 500 ms processing time on a GPU. This indicates a critical need for optimizing the segmentation approach to balance accuracy with performance, or alternatively, considering a trade-off that might slightly reduce accuracy for the sake of achieving real-time processing capabilities. Object Detection and Tracking are crucial components of our algorithm, with GPU-based processing time indicating the feasibility for real-time implementation. However, transitioning to CPU processing introduces substantial delays, particularly for Object Detection, which drastically exceeds the real-time requirement.

In summary, as shown in Table \ref{table:performance_metrics}, our framework exhibits the potential for real-time performance on systems with GPUs, achieving a processing time of 45.58 ms per frame, with 6.857 ms for safety evaluation. This capability necessitates a trade-off, specifically the sacrifice of segmentation smoothing to enhance location accuracy. Despite this compromise, performance on GPU-equipped systems indicates a promising direction. However, when operating on CPUs, the framework encounters significant slowdowns, particularly in object detection, pointing to the necessity of specialized hardware for optimal performance. Therefore, practical real-time application may necessitate the development of NPUs or similar technology that offers high computational power in a compact form factor, specifically tailored to streamline Object Detection tasks. Although this direction presents a promising avenue to ensure real-time applicability for predicted risk assessment, it falls outside the scope of our current research focus, highlighting an area for future exploration and development.

\begin{table}[ht]
\centering

\caption{Summary of Algorithm Time Costs by Component and Device}
\label{table:performance_metrics}
\resizebox{\textwidth}{!}{
\begin{tabular}{l|ccccc}
\hline
\textbf{Algorithm Category} & \textbf{Computer Vision [ms]} & \textbf{Safety Evaluation [ms]} & \textbf{Total Time Cost [ms]} & \textbf{Unit} & \textbf{Device} \\ \hline
\multirow{2}{*}{Risk Evaluation Framework} & 528.57 & 6.857 & 535.43 & Frame & GPU \\
                                            & \textgreater{}2s & 32.536 & \textgreater{}2s & Frame & CPU \\
\textbf{\multirow{2}{*}{\shortstack[l]{Risk Evaluation Framework \\ (Without Segmentation)}}} & \textbf{38.72} & \textbf{6.857} & \textbf{45.58} & \textbf{Frame} & \textbf{GPU} \\
                                                                  & 682.23 & 32.536 & 714.77 & Frame & CPU \\ \hline
\end{tabular}}
\end{table}

\section{Conclusion and future work}
\label{sec:section 4}
%what we propose
In this study, we proposed a comprehensive framework for pedestrian's potential risk evaluation using a novel surrogate safety measure, Predicted Post-Encroachment Time, derived through deep learning methods. 
% The core method
Our approach involves the deployment of deep learning models to accurately predict pedestrian arrival time, thus determining P-PET as a forward-looking measure of potential pedestrian risk.
% The Risk Evaluation Algorithm
The development of a potential risk evaluation algorithm based on the proposed framework uses P-PET to quantify the potential risk to pedestrians. To further refine our analysis, we classify pedestrians into distinct categories, kids, adults, and cyclists, and examine their impact within both the closer and the further area, thus enhancing the precision of our arrival time prediction models and the ensuing risk evaluation algorithm.

Our empirical investigation focused on a nonsignalized intersection adjacent to a kindergarten. 
By processing video data captured from an on-site CCTV camera, we extracted comprehensive trajectories for vehicles, kids, adults, and cyclists. A comparative analysis among three proposed deep learning models, along with a traditional historical average method, revealed optimal arrival time prediction models that varied by category of pedestrians and target location. Implementing our algorithm with calibrated thresholds and counters facilitated an effective assessment of pedestrian risk, resulting in a recall rate of 82.64\% and an F1 score of 78.09\% in evaluating potential risk. In particular, incorporating pedestrian classification into the evaluation framework markedly improved algorithm performance, especially in the further area, with adults showing greater predictability and kids presenting evaluation challenges due to their erratic movements. The cyclists showed moderate performance, attributed to their distinct crossing behavior. In summary, the results demonstrate the framework's ability to effectively identify potential risks through the use of P-PET, indicating its feasibility for real-time applications and its improved performance in risk evaluation across different categories of pedestrians.
%what result we can generate 

Our proposed framework will significantly benefit pedestrian safety through proactive measures, such as alerting pedestrians or activating vehicular brakes at crucial moments. Furthermore, the application of P-PET offers urban planners and policy makers greater insight into possible collision hot spots at intersections, guiding the strategic placement of traffic signs and the design of safer intersection layouts.
%Future work and limitation

However, our study is not without limitations. It focuses primarily on kids, adults, and cyclists, overlooking other vulnerable pedestrian groups, such as the elderly, disabled, and those under the influence of alcohol. Additionally, issues related to object overlap and fixed camera angles pose challenges in data extraction, which could be mitigated by integrating Lidar with video data. The pursuit of higher sample sizes and the exploration of real-time processing capabilities, particularly on edge devices, represent critical areas for future research. Additionally, the current safety indicator only considers temporal information and requires that spatial information is taken into account for risk evaluation. Although the current iteration of our algorithm does not meet the real-time processing threshold necessary for the application of edge devices, it lays a foundational blueprint for subsequent iterations of proactive pedestrian protection systems. As we continue to refine our approach, the evolution of hardware computing technologies along with the optimization of detection, tracking, and segmentation algorithms will be crucial to overcoming existing barriers to real-time processing. Our work serves as a stepping stone in the ongoing effort to proactively improve pedestrian safety at nonsignalized intersections.

\section{Acknowledgements}
This work was supported by Innovative Human Resource Development for Local Intellectualization program through the Institute of Information \& Communications Technology Planning \& Evaluation(IITP) grant funded by the Korea government(MSIT)(IITP-2024-00156287)

% \newpage

% \bibliographystyle{trb}
% \bibliography{trb_template}
\newpage
\printcredits

\bibliographystyle{unsrtnat}
\bibliography{trb_template}

\end{document}